\documentclass[nohyperref]{article}

\usepackage{microtype}
\usepackage{graphicx}
\usepackage{subfigure}
\usepackage{booktabs} 

\usepackage{hyperref}
\usepackage{multirow}

\usepackage[accepted]{icml2022}

\usepackage{amsmath}
\usepackage{amssymb}
\usepackage{mathtools}
\usepackage{amsthm}
\usepackage{newtxmath}

\usepackage[capitalize,noabbrev]{cleveref}

\theoremstyle{plain}

\theoremstyle{definition}

\theoremstyle{remark}

\usepackage[textsize=tiny]{todonotes}

\icmltitlerunning{Understanding Robust Overfitting of Adversarial Training and Beyond}

\begin{document}

\twocolumn[
\icmltitle{Understanding Robust Overfitting of Adversarial Training and Beyond}




\icmlsetsymbol{intern}{\dag}

\begin{icmlauthorlist}
\icmlauthor{Chaojian Yu}{sch1,intern}
\icmlauthor{Bo Han}{sch2}
\icmlauthor{Li Shen}{comp}
\icmlauthor{Jun Yu}{sch3}
\icmlauthor{Chen Gong}{sch4}
\icmlauthor{Mingming Gong}{sch5}
\icmlauthor{Tongliang Liu}{sch1}
\end{icmlauthorlist}

\icmlaffiliation{sch1}{TML Lab, Sydney AI Centre, The University of Sydney, Sydney, Australia}
\icmlaffiliation{sch2}{Department of Computer Science, Hong Kong Baptist University, Hong Kong, China}
\icmlaffiliation{comp}{JD Explore Academy, Beijing, China}
\icmlaffiliation{sch3}{Department of Automation, University of Science and Technology of China, Hefei, China}
\icmlaffiliation{sch4}{School of Computer Science and Engineering, Nanjing University of Science and Technology, Nanjing, China}
\icmlaffiliation{sch5}{School of Mathematics and Statistics, The University of Melbourne, Melbourne, Australia}

\icmlcorrespondingauthor{Tongliang Liu}{tongliang.liu@sydney.edu.au}

\icmlkeywords{Adversarial Training, Robust overfitting, Machine Learning, ICML}

\vskip 0.3in
]



\printAffiliationsAndNotice{$^{\dag}$ This work is done during an internship at JD Explore Academy.}  

\begin{abstract}
Robust overfitting widely exists in adversarial training of deep networks. The exact underlying reasons for this are still not completely understood. Here, we explore the causes of robust overfitting by comparing the data distribution of \emph{non-overfit} (weak adversary) and \emph{overfitted} (strong adversary) adversarial training, and observe that the distribution of the adversarial data generated by weak adversary mainly contain small-loss data. However, the adversarial data generated by strong adversary is more diversely distributed on the large-loss data and the small-loss data. Given these observations, we further designed data ablation adversarial training and identify that some small-loss data which are not worthy of the adversary strength cause robust overfitting in the strong adversary mode. To relieve this issue, we propose \emph{minimum loss constrained adversarial training} (MLCAT): in a minibatch, we learn large-loss data as usual, and adopt additional measures to increase the loss of the small-loss data. Technically, MLCAT hinders data fitting when they become easy to learn to prevent robust overfitting; philosophically, MLCAT reflects the spirit of turning waste into treasure and making the best use of each adversarial data; algorithmically, we designed two realizations of MLCAT, and extensive experiments demonstrate that MLCAT can eliminate robust overfitting and further boost adversarial robustness.
\end{abstract}

\section{Introduction}
\label{Introduction}
Adversarial examples easily mislead deep neural networks (DNNs) to produce incorrect outputs, which raises security concerns in various real-world applications since adversarial noise is usually small and human-imperceptible~\citep{szegedy2013intriguing,goodfellow2014explaining}. The vulnerability of DNNs has attracted extensive attention and led to a large number of defense techniques. Across existing defenses, adversarial training (AT)~\citep{goodfellow2014explaining,madry2017towards} is one of the strongest empirical defenses~\citep{athalye2018obfuscated}. AT incorporates adversarial examples into the training process and can be viewed as solving a min-max optimization problem~\citep{madry2017towards}.

Unfortunately, robust overfitting seems inevitable in adversarial training of deep networks: after a certain point in AT, i.e., shortly after the first learning rate decay, the robust performance on test data will continue to degrade with further training~\citep{rice2020overfitting}. More pessimistically, conventional remedies for overfitting in deep learning, including explicit  regularizations, data augmentation, etc., cannot gain improvements upon early stopping~\citep{rice2020overfitting}. Nevertheless, early stopping might not be our desideratum since the expectation of double descent phenomena occurred in adversarial training ~\citep{nakkiran2019deep}.

Robust overfitting widely exists in adversarial training of deep networks, and the exact underlying reasons for this are still not completely understood. To explore the causes of robust overfitting, we compare the data distribution of \emph{non-overfit} adversarial training (weak adversary) and \emph{overfitted} adversarial training (strong adversary). We observed that the training data of \emph{non-overfit} adversarial training mainly contains small-loss data; while the distribution of training data of \emph{overfitted} adversarial training is more divergent, usually mixed with a considerable proportion of small-loss data and large-loss data. Given these observations, we conduct a range of data ablation adversarial training experiments. By removing training data from various loss ranges, we study the impact of small-loss data and large-loss data on robust overfitting. Our results show that robust overfitting is actually caused by some small-loss data in \emph{overfitted} adversarial training. Adversarial data that are not worthy of the adversary strength make adversarial training worse, which might be explained by the fact the network becomes more robust as the adversarial training progresses, making some generated adversarial data relatively less aggressive, and when their loss drops to a certain level, these adversarial data eventually lead to robust overfitting.

To relieve this issue of robust overfitting due to the small-loss data, we propose \emph{minimum loss constrained adversarial training} (MLCAT). Specifically, MLCAT works in each mini-batch: it learns large-loss data as usual, and if there are any small-loss data, it implements additional measures on these data to increase their loss. It is a general adversarial training prototype, where small-loss data and large-loss data can be separated by a threshold. The implementation of additional measures to increase data loss can be versatile. For instance, we designed two representative methods for the realizations of MLCAT: \emph{loss scaling} and \emph{weight perturbation}. They adopt different strategies to improve data loss, e.g., manipulating the learning rate or model parameters, respectively. Extensive experiments show that they not only eliminate robust overfitting, but also further boost adversarial robustness.

MLCAT can be justified as follows. It is of vital importance to distinguish small-loss data and large-loss data in the context of adversarial training. The inner maximization in min-max optimization is to generate worst-case adversarial example that maximizes the classification loss ~\citep{madry2017towards,wang2019convergence}. However, the adversarial data generated by the adversary are not always qualified since the model become more robust during training. In strong adversary mode, there are some data, even if they are attacked by adversary, they are still easy to be fitted to the network, which not only fails to enhance the adversarial robustness, but also leads to robust overfitting. Therefore, if the network can be trained on the data with a minimum loss constrained, then robust overfitting may not occur. Technically, MLCAT is a specially designed adversarial training prototype to hinder data fitting when they become easy to learn, which provides a novel viewpoint on the adversary in adversarial training.

Furthermore, it is known that the sample complexity of robust generalization in adversarial training is significantly larger than that of standard generalization in natural training ~\citep{schmidt2018adversarially}. A substantially larger dataset is required to achieve the double descent phenomena in AT ~\citep{carmon2019unlabeled,uesato2019labels,zhai2019adversarially}. However, the cost of collecting and training additional data should not be neglected. Unlike them, we delve into the causes of robust overfitting and aim to eliminate robust overfitting without exploiting additional training data. Nevertheless, simply removing small-loss data which causes robust overfitting might not be a good choice, due to the benefit of sample size~\citep{schmidt2018adversarially}. Therefore, we adopt additional measures to increase their loss to take full advantage of each adversarial data. In this sense, philosophically, MLCAT reflects the spirit of turning waste into treasure and making the best use of each adversarial data.

\section{Related Work}
\label{Related Work}
This section briefly reviews relevant adversarial learning methods from two perspectives: adversarial training and robust overfitting.

\subsection{Adversarial Training}
Adversarial training (AT) has been demonstrated to be the most effective method for defending against adversarial attacks \cite{athalye2018obfuscated,croce2022evaluating}. Let $\mathcal{X}$ and $\mathcal{Y}$ be the input and output domains. Let $\mathcal{D}=\{(x_i,y_i)\}^{n}_{i=1}$ be the training dataset with n samples, and $\mathcal{B}_{\epsilon}^{p}(x_i) = \{x_i' \in \mathcal{X}:||x_i'-x_i||_p \le \epsilon\}$ be their adversarial regions, where $\epsilon$ is the maximum perturbation constraint. In AT, the training data are all sampled from adversarial regions. \citet{madry2017towards} formulated AT as a min-max optimization problem:
\begin{equation}
\min_w \sum_{i} \max_{x_{i}' \in \mathcal{B}_{\epsilon}^{p}(x_i)} \ell(f_w(x_{i}'),y_{i}),
\label{min-max}
\end{equation}
where $f_w$ is the DNN classifier with weight $w$, and $\ell(\cdot)$ is the loss function. The inner maximization pass is to find adversarial example $x_{i}'$ that maximizes the loss. The outer minimization pass is to optimize network parameters $w$ that minimize the loss on adversarial examples. The commonly used technique to solve the inner maximization problem is Projected Gradient Descent (PGD) \cite{madry2017towards}, which perturbs normal example $x_i$ for multiple step $K$ with step size $\alpha$:
\begin{equation}
\label{pgd}
    x_i^k=\Pi_{\mathcal{B}_{\epsilon}^{p}(x_i)}(x_i^{k-1}+\alpha \cdot \mathrm{sign}(\nabla_{x_i^{k-1}}\ell(f_w(x_i^{k-1}),y_i))),
\end{equation}
where 
$x_i^k$ denotes the adversarial example at step $k$, and $\Pi_{\mathcal{B}_{\epsilon}^{p}(x_i)}$ is the projection operator.

Another typical AT variant is TRADES \citep{zhang2019theoretically}, which optimizes a regularized surrogate loss that is a tradeoff between the natural accuracy and adversarial robustness:
\begin{equation}
  \begin{aligned}
    \min_w \sum_{i} \big\{ & \mathrm{CE}(f_{w}(x_i),y_i) \\
    & + \beta \cdot \max_{x_i' \in \mathcal{B}_{\epsilon}^{p}(x_i)} \mathrm{KL}(f_{w}(x_i)||f_{w}(x_i')) \big\},
  \end{aligned}
\end{equation}
where CE is the cross-entropy loss that encourages the network to maximize the natural accuracy, KL is the Kullback-Leibler divergence that encourages to improve the robust accuracy, and $\beta$ is the hyperparameter to control the tradeoff between natural accuracy and adversarial robustness.

Based on AT, subsequent works have been developed to further improve its performance, such as adversarial regularization~\citep{kannan2018adversarial,wang2019improving}, curriculum learning~\citep{cai2018curriculum,wang2019convergence,zhang2020attacks}, input denoising~\citep{wu2021attacking}, modeling adversarial noise~\citep{zhou2021modeling,xia2019anchor,xia2020part}, feature alignments~\citep{yan2021cifs,bai2021improving} and AT tricks~\citep{pang2020bag,gowal2020uncovering}.

\subsection{Robust Overfitting}
Robust overfitting widely exists in AT, but the cause of robust overfitting remains unclear~\citep{rice2020overfitting}. \citet{schmidt2018adversarially} theoretically proposes that a substantially large dataset is required to achieve robust generalization, which is supported by empirical results in derivative works, such as AT with semi-supervised learning \citep{carmon2019unlabeled,uesato2019labels,zhai2019adversarially}, robust local feature \citep{song2019robust} and data interpolation \citep{zhang2019adversarial,lee2020adversarial,chen2021guided}. Different from these works, we eliminate robust overfitting without exploiting more training data. Separate works have also attempt to mitigate robust overfitting by sample reweighting \citep{wang2019improving,zhang2020geometry}, weight perturbation \citep{wu2020adversarial,yu2022robust} and weight smoothing \citep{chen2020robust}. Although robust overfitting has been widely investigated, there still lacks an explanation of why it occurs. \citep{dong2022exploring} explores the memorization effect in AT and thinks large-loss data induces robust overfitting. In contrast, we delve into the causes of robust overfitting by investigating the data distribution of AT and identify that some small-loss data caused robust overfitting in the strong adversary mode. Then, the proposed MLCAT prototype explicitly adjusts the training data based on their loss to avoid robust overfitting. Finally, experiments on two specific realizations of MLCAT demonstrate the effectiveness on eliminating robust overfitting and robustness improvement.

\begin{figure*}[t]
\centering
    \subfigure[]{
        \includegraphics[width=0.48\columnwidth]{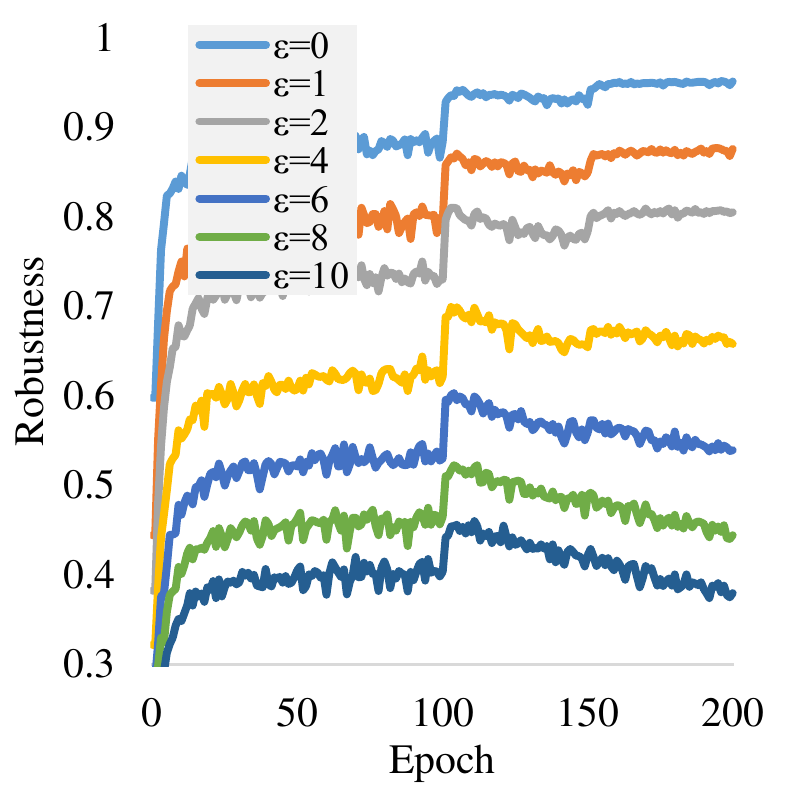}
    }
    \subfigure[Data distribution under perturbation size of 0, 1, and 2 (from left to right)]{
        \includegraphics[width=0.48\columnwidth]{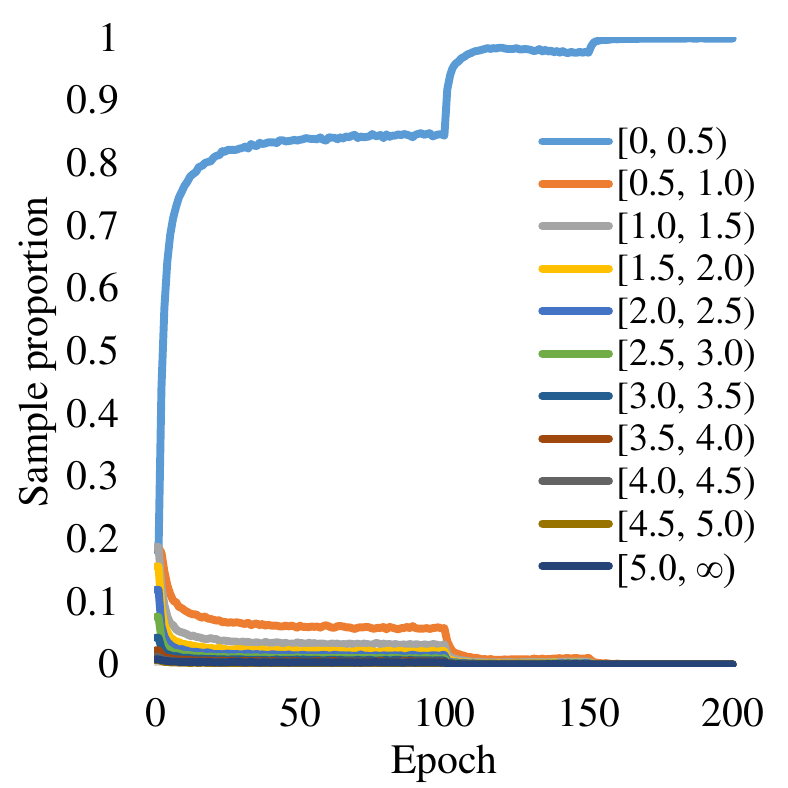}
        \includegraphics[width=0.48\columnwidth]{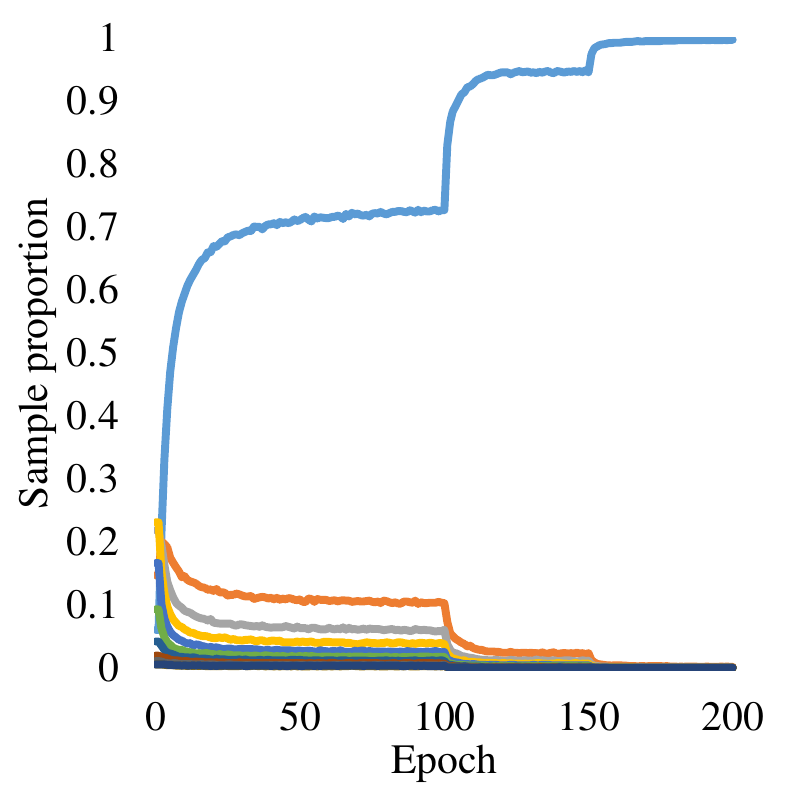}
        \includegraphics[width=0.48\columnwidth]{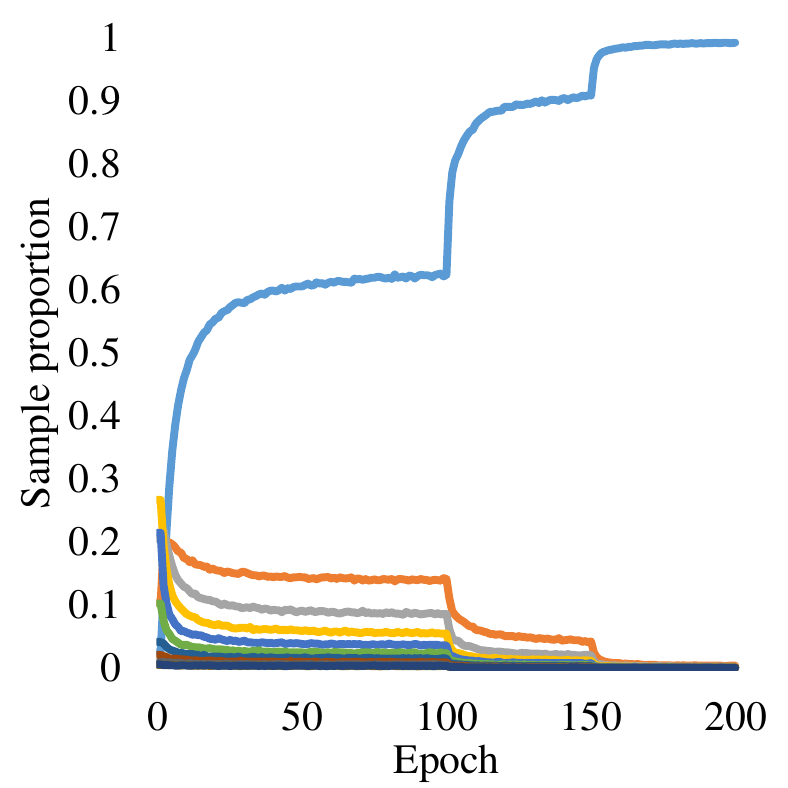}
    }
    \\
    \subfigure[Data distribution under perturbation size of 4, 6, 8, and 10 (from left to right)]{
        \includegraphics[width=0.48\columnwidth]{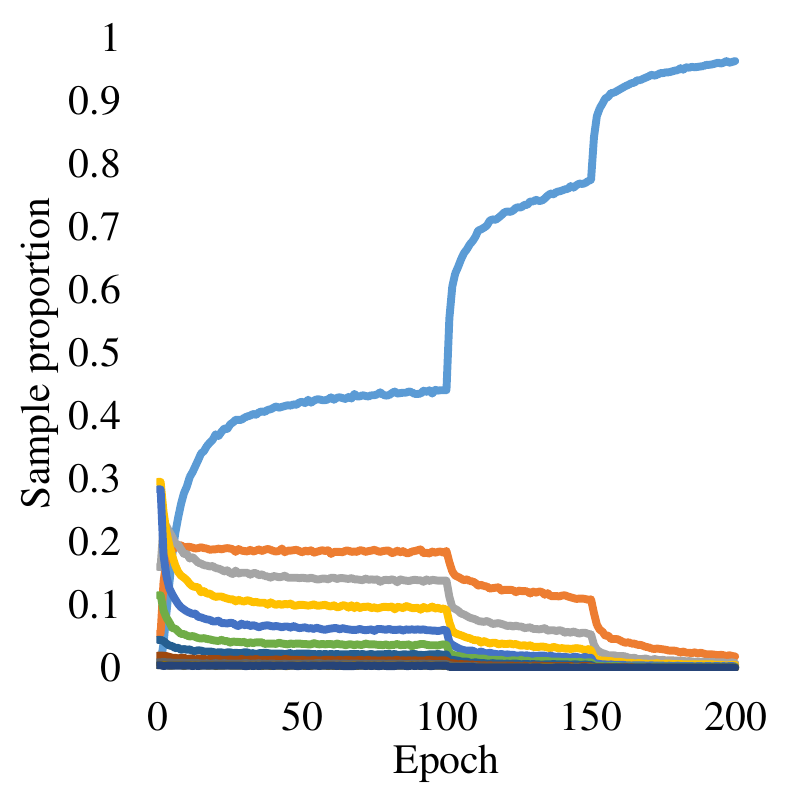}
        \includegraphics[width=0.48\columnwidth]{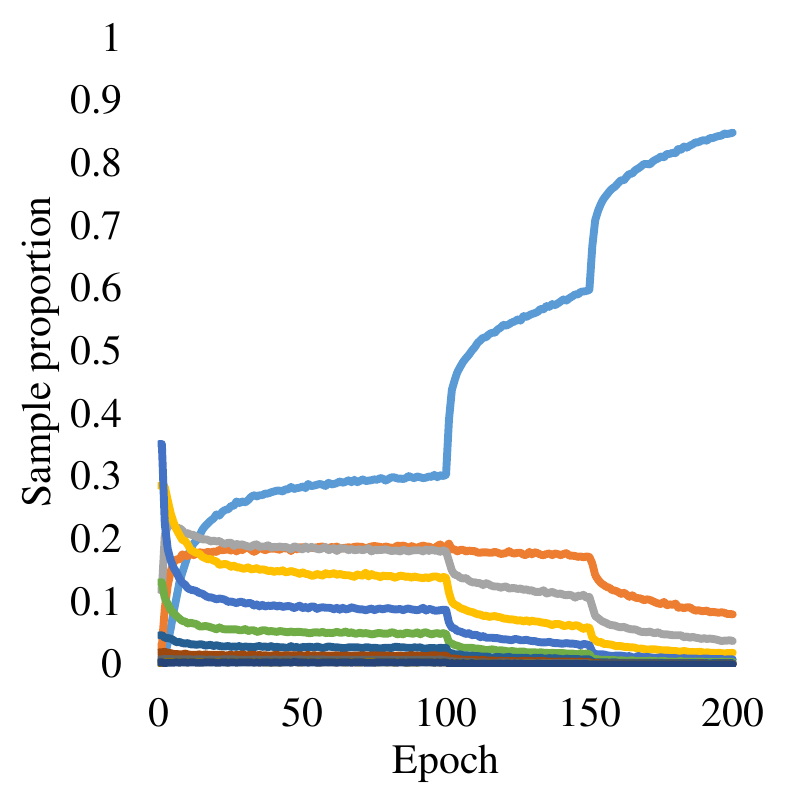}
        \includegraphics[width=0.48\columnwidth]{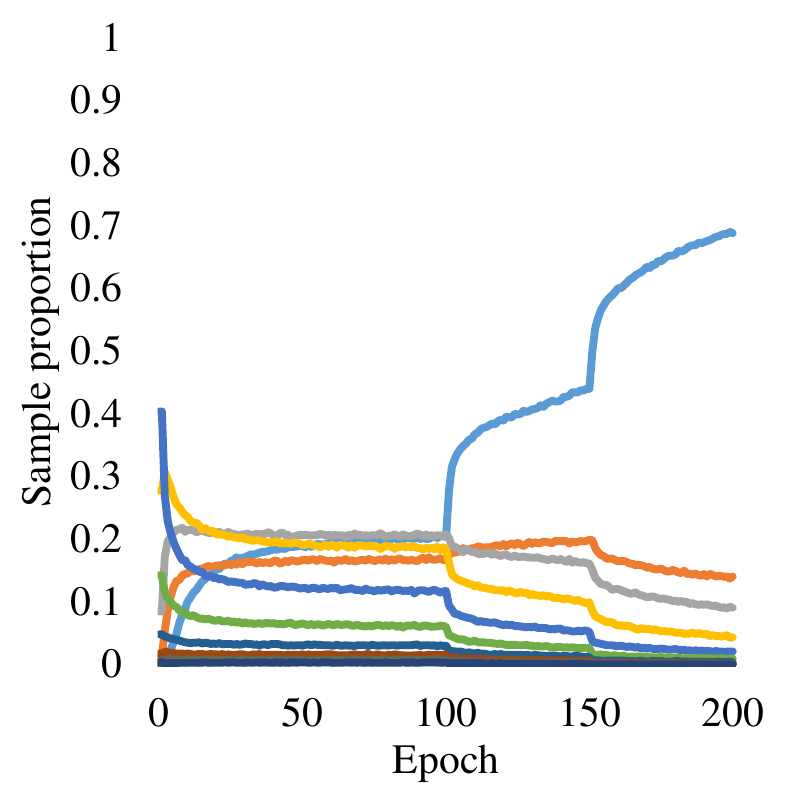}
        \includegraphics[width=0.48\columnwidth]{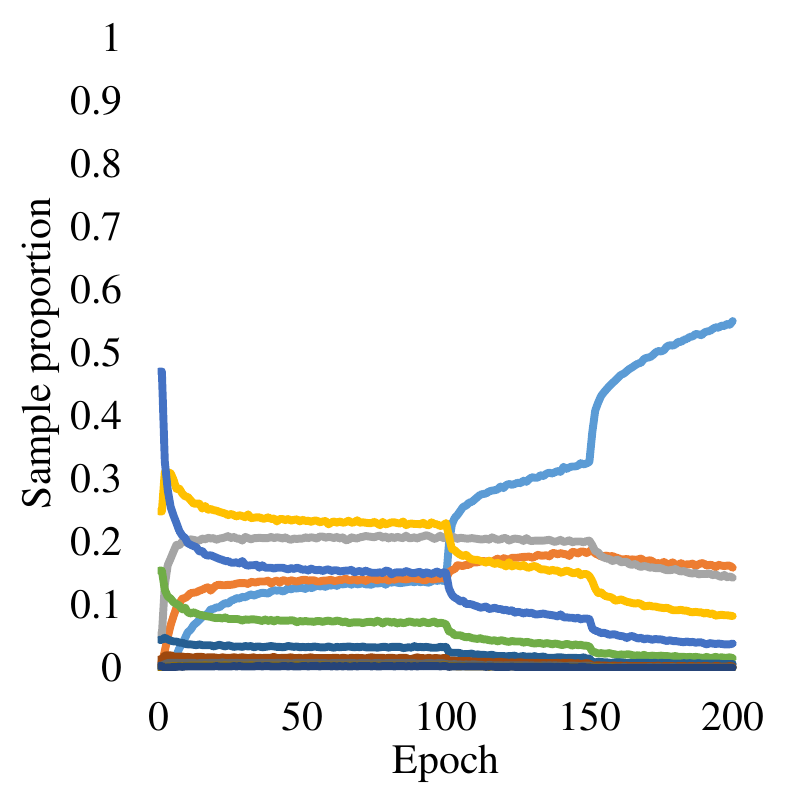}
    }
\caption{(a): The test robustness of adversarial training with various perturbation size $\epsilon$; (b) and (c): The distribution of training data in different loss ranges under various perturbation size $\epsilon$.}
\label{fig:1}
\end{figure*}

\begin{figure*}[t]
\centering
    \subfigure[]{
        \includegraphics[width=0.64\columnwidth]{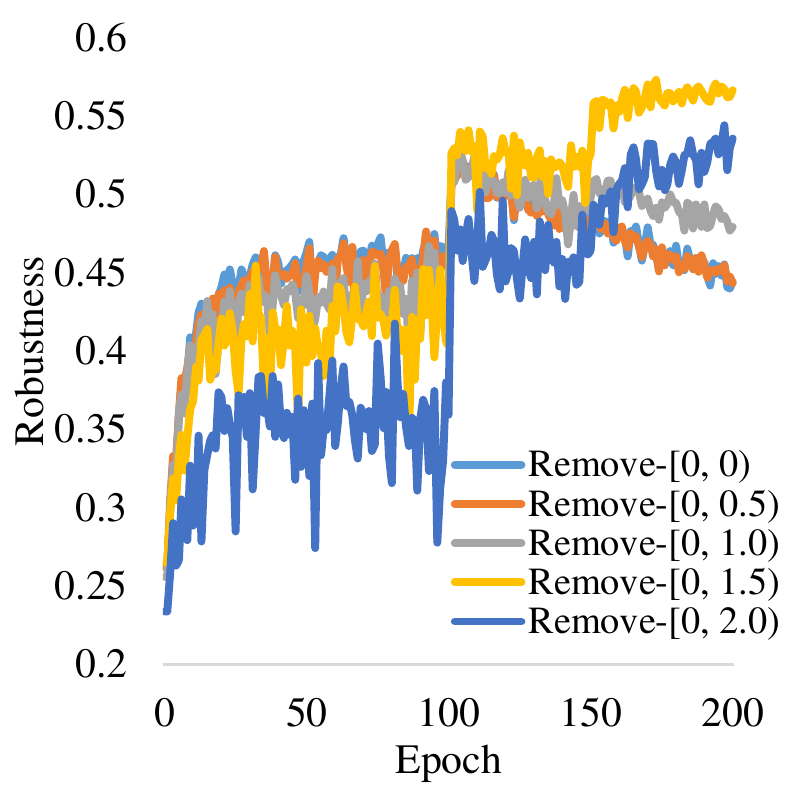}
    }
    \subfigure[]{
        \includegraphics[width=0.64\columnwidth]{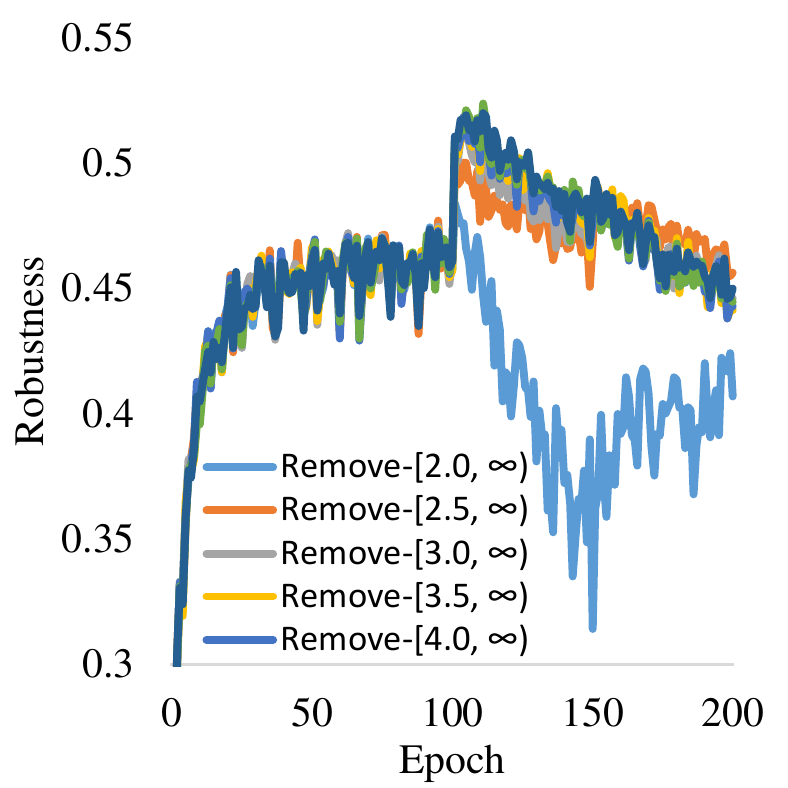}
    }
    \subfigure[]{
        \includegraphics[width=0.64\columnwidth]{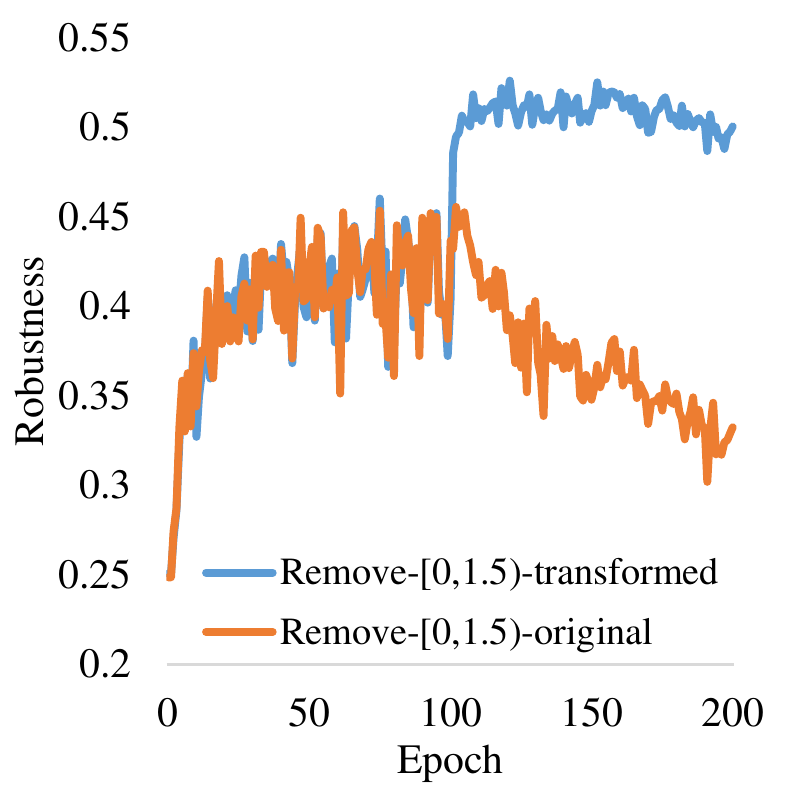}
    }
\caption{The test robustness of adversarial training (a) without small-loss data in various loss ranges; (b) without large-loss data in various loss ranges, and (c) without original small-loss data and transformed small-loss data.}
\label{fig:2}
\end{figure*}

\section{Understanding Robust Overfitting in AT}
\label{Understanding Robust Overfitting}
In this section, we first study the robust overfitting by comparing the data distribution of non-overfit (weak adversary) and overfitted (strong adversary) adversarial training (Section \ref{3-1}). Based on our observations, we further propose data ablation adversarial training to identify the specific causes of robust overfitting (Section \ref{3-2}). Finally, we develop a novel adversarial training prototype, MLCAT, to eliminate robust overfitting (Section \ref{3-3}).

\subsection{Non-overfit AT vs. Overfitted AT: Data Distribution Perspective}
\label{3-1}
We compare the data distribution of non-overfit (weak adversary) and overfitted (strong adversary) adversarial training. Specifically, we change the strength of the adversary by adjusting the maximum perturbation size $\epsilon$. We train PreAct ResNet-18 on CIFAR10 under $L_\infty$ threat model using various $\epsilon$ from 0, 1, 2, 4, 6, 8 to 10. In each setting, we evaluate the accuracy of trained model on CIFAR10 test data which are attacked with the same $\epsilon$. The test robustness of adversarial training with different adversary is shown in Figure \ref{fig:1}(a). We can observe that there is no robust overfitting in the case of weak adversary ($\epsilon$ is small). However, in the case of strong adversary ($\epsilon$ is large), robust overfitting is a dominant phenomenon. For each case, we then visualize the distribution of training data in different loss ranges, which are shown in Figure \ref{fig:1}(b) and Figure \ref{fig:1}(c).

From Figure \ref{fig:1}(b) and Figure \ref{fig:1}(c), we can observe that the data distribution of overfitted adversarial training is obviously mismatched with that of the non-overfit adversarial training: the training data of non-overfit adversarial training mainly contains small-loss data. In contrast, the data distribution of overfitted adversarial training is more divergent, usually containing a considerable proportion of small-loss data and large-loss data.
Given this observation, we wonder (Q1): \emph{if we suppress the large-loss data in overfitted adversarial training to align the data distribution of non-overfit adversarial training, will it eliminate robust overfitting?}

On the other hand, robust overfitting occurs when the adversary becomes stronger. For strong adversary, the large-loss data is expected. In other words, for adversarial training with strong adversary, these large-loss data are good data, while these small-loss data may be bad data. From this perspective, we may ask another question (Q2): \emph{if we suppress the small-loss data in overfitted adversarial training that does not match the strength of adversary, will it eliminate robust overfitting?}

It is worth noting that the robust overfitting behaviors and their data distribution can commonly be observed across a variety of datasets, model architectures, and threat models (shown in Appendix \ref{AP_A}), indicating that it is a general phenomenon in adversarial training. Considering the two questions above, we conduct further analysis in the next subsection to identify the specific causes of robust overfitting.

\subsection{Causes of Robust Overfitting}
\label{3-2}
To answer the two questions in Section \ref{3-1}, we conduct adversarial training with fixed perturbation size ($\epsilon=8$) in a data ablation manner. Specifically, we train PreAct ResNet-18 on CIFAR10 under $L_\infty$ threat model by removing training data from various loss ranges. To answer Q1, we remove the large-loss data within the specified loss range before robust overfitting occurs (for example, at 100th epoch), which is for the stability of optimization. To answer Q2, we remove small-loss data within the specified loss range from the beginning of training. As shown in Figure \ref{fig:2}(a) and Figure \ref{fig:2}(b), we can observe that adversarial training without large-loss data still has a significant robust overfitting phenomenon, which indicates the strategy of aligning with the data distribution of non-overfit adversarial training is invalid. In contrast, adversarial training without small-loss data can eliminate robust overfitting, which indicates that adversarial data which is not match the strength of adversary make adversarial training worse. Notice that similar experimental results can be obtained across a variety of datasets, model architectures, and threat models (shown in Appendix \ref{AP_B}). These empirical results clearly verified that the small-loss data causes robust overfitting in strong adversary mode.

It is worth noting that the small-loss data contains two sources. As shown in Figure \ref{fig:1}(b) and Figure \ref{fig:1}(c), one is the original data before the learning rate decay (before 100th epoch), and the other is transformed from other loss ranges (after 100th epoch). We further investigate the respective effects on robust overfitting of the two parts of small-loss data, as shown in Figure \ref{fig:2}(c). It is observed that robust overfitting is mainly due to these transformed small-loss data, which might be explained by the fact that as adversarial training progresses, the network becomes more robust, so some generated adversarial data are relatively less aggressive, and when they weaken to a certain extent, these adversarial data eventually lead to robust overfitting.
In this paper, we take measures to eliminate robust overfitting based on general data loss. Thus we make no distinction between them and refer to them together as small-loss data.
Given existing results, it does not seem to be a wise choice to remove the small-loss data directly in order to eliminate robust overfitting, because this will reduce the training sample size. In the next subsection, we introduce a novel adversarial training prototype to address this issue.

\subsection{A Prototype of MLCAT}
\label{3-3}
As mentioned in Section \ref{3-2}, the small-loss data causes robust overfitting in adversarial training with strong adversary. To relieve this issue, we propose to train network on adversarial data under a minimum loss constraint, dubbed as \emph{minimum loss constrained adversarial training} (MLCAT). We adopt additional measures to increase the loss of small-loss data, so as to ensure that there is neither robust overfitting nor sample size decline.

For now, let us omit the technical details, and assume that we have a base AT method that is implemented as an algorithm with a \emph{inner maximization pass} and a \emph{outer minimization pass}. The inner maximization pass generates adversarial data that maximizes the loss, and then the outer minimization pass returns the gradient by backward propagating the average loss. With this algorithmic abstraction, we can present MLCAT at a high level.

The MLCAT prototype is given in Algorithm \ref{alg:1}. Since it is only a prototype, it serves as a versatile approach where the loss adjustment strategy $\mathcal{S}$ and minimum loss condition $\ell_{min}$ can be flexibly implemented depend on the base AT algorithm $\mathcal{A}$. Algorithm \ref{alg:1} runs as follows. Given the mini-batch $\mathcal{B}$, the inner maximization pass of $\mathcal{A}$ is called in Line 2. Then the loss values are manipulated in Line 3-13 where they are reduced to a scalar for backpropagation. Before the for loop, the loss accumulator $\ell_{\mathcal{B'}}$ is initialized in Line 4. Subsequently, in Line 7, $\ell_i$ is added to $\ell_{\mathcal{B'}}$ if $x_i$ is regarded as large-loss data, which will result in normal gradient descent by $\mathfrak{O}$ in Line 15; in Line 9-10, $\ell_i^{\mathcal{S}}$ is adjusted by strategy $\mathcal{S}$ and then added to $\ell_{\mathcal{B'}}$ if $x_i$ is regarded as small-loss data, which will lead to adjusted gradient descent by $\mathfrak{O}$ in Line 15. After the for loop, the accumulated loss $\ell_{\mathcal{B'}}$ is divide by the batch size $m$ in Line 13 to obtain the average loss. Finally, the outer minimization pass of $\mathcal{A}$ is called in Line 14, and the optimizer $\mathfrak{O}$ updates the model $f_w$ in Line 15.

Last but not least, notice that MLCAT is extremely general. MLCAT becomes standard AT if $\mathcal{S}$ is identical mapping; it becomes data ablation adversarial training if $\mathcal{S}$ always returns 0. In AT, the sample size plays an important role in robust generalization~\citep{schmidt2018adversarially}. Hence, we use $\mathcal{S}$ to convert the small-loss data into the large-loss data. This converting behavior makes the network learn all the training data without robust overfitting, which is the key of MLCAT and what we meant by turning waste into treasure in the \emph{abstract}. Moreover, the hyperparameter $\ell_{min}$ controls the range of small-loss data: when $\ell_{min} \le 0$, it becomes standard adversarial training again; when $\ell_{min} \rightarrow \infty$, all data will be adjusted by strategy $\mathcal{S}$. Nevertheless, forcibly adjusting the loss of all data will inevitably have negative impacts on network optimization. In MLCAT, $\ell_{min}$ is closely related to the perturbation size $\epsilon$ and specific tasks. For example, when $\epsilon$ is small, the degradation of adversarial attack is very limited, thus $\ell_{min}$ may not needed since robust overfitting does not occur. Consequently, $\ell_{min}$ should be carefully tuned in practice in consideration of the elimination of robust overfitting as well as robustness improvement.

\begin{algorithm}[t]
\caption{MLCAT-prototype (in a mini-batch).}
\label{alg:1}
{\bf Require:} base adversarial training algorithm $\mathcal{A}$, optimizer $\mathfrak{O}$, network $f_w$, training data $\mathcal{D}=\{(x_i,y_i)\}^{n}_{i=1}$, mini-batch $\mathcal{B}$, batch size $m$, minimum loss conditions $\ell_{min}$ for $\mathcal{A} $, loss adjustment strategy $\mathcal{S}$
\begin{algorithmic}[1]
   \STATE Sample a mini-batch $\mathcal{B}=\{(x_i,y_i)\}^{m}_{i=1}$ from $\mathcal{D}$
   \STATE $\mathcal{B'} = \mathcal{A}.\mathrm{inner\_maximization}(f_w,\mathcal{B})$ 
   \STATE $\{\ell_i\}^{m}_{i=1} \leftarrow \ell(f_w,\mathcal{B'})$
   \STATE $\ell_{\mathcal{B'}} \leftarrow 0$ \hfill \# initialize loss accumulator
   \FOR{$i=1,...,m$}
   \IF{$\ell_i \ge \ell_{min}$}
   \STATE $\ell_{\mathcal{B'}} = \ell_{\mathcal{B'}} + \ell_i$ 
   \ELSE 
   \STATE $\ell_i^{\mathcal{S}} \leftarrow \mathcal{S}(f_w,x_i',\ell_{min})$ \hfill \# adjust loss
   \STATE $\ell_{\mathcal{B'}} = \ell_{\mathcal{B'}} + \ell_i^{\mathcal{S}}$ \hfill \# accumulate adjusted loss
   \ENDIF
   \ENDFOR
   \STATE $\ell_{\mathcal{B'}} \leftarrow \ell_{\mathcal{B'}}/m$ \hfill \# average accumulated loss
   \STATE $\nabla_{w} \leftarrow \mathcal{A}.\mathrm{outer\_minimization}(f_w,\ell_{\mathcal{B'}})$
   \STATE $\mathfrak{O}.\mathrm{step}(\nabla_{w})$
\end{algorithmic}
\end{algorithm}

\section{Two Realizations of MLCAT}
\label{Two Realizations}
In this section, we illustrate what $\mathcal{S}$ can be used in MLCAT. We realize MLCAT through \emph{loss scaling} that belongs to the loss-correction approach and \emph{weight perturbation} that belongs to the parameter-correction approach. Loss scaling and weight perturbation are two representative and more importantly orthogonal methods, which spotlights the great versatility of MLCAT.

\textbf{MLCAT through loss scaling.} The loss-correction approach creates a corrected loss from original loss $\ell_i$ and then trains the model $f_w$ based on the corrected loss. Misclassification-aware \cite{wang2019improving} is the most primitive method in this direction. It introduces a regularizer to enhance the effect of misclassified examples on the final robustness of adversarial training.

In order to increase the loss of small-loss data. We adopt a straightforward scaling technique to correct the loss of small-loss data. Since we aim to eliminate robust overfitting, we leverage the minimum loss conditions $\ell_{min}$ as a constraint and scale the loss $\ell_i$ as follows:
\begin{equation}
\label{1-increase}
\ell_i^{\mathcal{S}} = \frac{\ell_{min}}{\ell_i} \cdot \ell_i = \ell_{min},
\end{equation}
where $\frac{\ell_{min}}{\ell_i}$ is the scaling coefficient, which is always greater than 1. The smaller the original loss $\ell_i$, the larger the scaling coefficient, and vice versa. Previous works \citep{zhang2020geometry,hitaj2021evaluating} show that scaling loss makes the trained model more sensitive to the logit-scaling attack. It does not matter since our aim is to verify the cause of robust overfitting, and the realization of MLCAT based on loss scaling is just designed for this purpose. Note that there is no difference between scaling losses and scaling gradients, since scaling coefficient has the same effects in increasing losses and increasing the learning rate inside $\mathfrak{O}$. Therefore, loss scaling can be regarded as learning small-loss data with a larger learning rate to effectively prevent the network from fitting these data. We refer the implementation method based on loss scaling as MLCAT$_{\mathrm{LS}}$.

\textbf{MLCAT through weight perturbation.} On the other hand, the parameter-correction approach generates perturbation to the model weights, and trains the network on the perturbative parameters. AWP \citep{wu2020adversarial} is the most primitive method in this direction. It develops a double-perturbation mechanism that adversarially perturbs both inputs and weights to reduce the robust generalization gap:
\begin{equation}
\min_w \max_{v \in V} \sum_i \max_{x_{i}' \in \mathcal{B}_{\epsilon}^{p}(x_i)} \ell (f_{w+v}(x_{i}'),y_{i}),
\label{AWP}
\end{equation}
where $v$ is the adversarial weight perturbation, which is generated by maximizing the classification loss:
\begin{equation}
v = \nabla_{w}\sum_{i} \ell_i.
\label{AWP}
\end{equation}
In order to increase the loss of small-loss data. We adopt the weight perturbation technique to generate perturbation noise for the small-loss data in a targeted manner. Similarly, since we aim to eliminate robust overfitting, we leverage the minimum loss conditions $\ell_{min}$ as a constraint and generate the perturbation noise $v$ as follows:
\begin{equation}
v = \nabla_{w}\sum_{i} \vmathbb{1}(\ell_i \le \ell_{min})~\ell_i,
\label{WP}
\end{equation}
where $\vmathbb{1}(\ell_i \le \ell_{min})$ is an indicator function, which will output 1 if $\ell_i \le \ell_{min}$ and 0 if $\ell_i > \ell_{min}$. After obtaining the perturbation noise $v$, we scale the perturbation noise according to the norm of $w$ to get the final weight perturbation $v = \gamma\frac{||w||}{||v||}v$, where $\gamma$ is the weight perturbation size. Then, the adjusted loss of small-loss data can be expressed as:
\begin{equation}
\label{2-increase}
\ell_i^{\mathcal{S}} = \ell (f_{w+v}(x_{i}'),y_{i}).
\end{equation}
Note that although weight perturbation technique can increase the loss of these small-loss data, it can not guarantee that they all satisfy the minimum loss condition $\ell_{min}$, as the goal of Eq.(\ref{WP}) is to maximize the overall loss of the small-loss data in the entire mini-batch rather than instance level. Anyway, the perturbation noise generated by Eq.(\ref{WP}) will effectively prevent the network from fitting these small-loss data. Therefore, \emph{weight perturbation} can be regarded as preventing robust overfitting by manipulating the model parameters, which is different from \emph{loss scaling} and is orthogonal in implementation. We refer the implementation method based on weight perturbation as MLCAT$_{\mathrm{WP}}$.

\begin{table*}[t]
\small
  \caption{Test robustness (\%) on CIFAR10. We omit the standard deviations of 5 runs as they are very small ($< 0.6\%$).}
  \label{table:1}
  \centering
  \begin{tabular}{cclccccccc}
    \toprule
    \multirow{2}*{Network} & \multirow{2}*{Threat Model} & \multirow{2}*{Method} & \multicolumn{3}{c}{PGD-20} & & \multicolumn{3}{c}{AA} \\
    \cmidrule{4-6}
    \cmidrule{8-10}
    & & & Best & Last & Diff & & Best & Last & Diff \\
    \midrule
    \multirow{6}*{PreAct ResNet-18} & \multirow{3}*{$L_\infty$} & AT & 52.29 & 44.43 & -7.86 & & 47.99 & 42.08 & -5.91 \\
     & & MLCAT$_{\mathrm{LS}}$ & 56.90 & 56.87 & \textbf{-0.03} & & 28.12 & 26.93 & -1.19\\
     & & MLCAT$_{\mathrm{WP}}$ & \textbf{58.48} & \textbf{57.65} & -0.83 & & \textbf{50.70} & \textbf{50.32} & \textbf{-0.38} \\
    \cmidrule{2-10}
    & \multirow{3}*{$L_2$} & AT & 69.27 & 65.86 & -3.41 & & 67.70 & 64.64 & -3.06\\
    & & MLCAT$_{\mathrm{LS}}$ & 73.16 & 72.48 & -0.68 & & 49.7 & 48.94 & -0.76\\
    & & MLCAT$_{\mathrm{WP}}$ & \textbf{74.38} & \textbf{73.86} & \textbf{-0.52} & & \textbf{70.46} & \textbf{70.15} & \textbf{-0.31} \\
     \midrule
    \multirow{6}*{Wide ResNet-34-10} & \multirow{3}*{$L_\infty$} & AT & 55.57 & 47.37 & -8.20 & & 52.13 & 46.09 & -6.04\\
    & & MLCAT$_{\mathrm{LS}}$ & \textbf{64.73} & \textbf{63.94} & -0.79 & & 35.00 & 34.51 & -0.49\\
    & & MLCAT$_{\mathrm{WP}}$ & 62.50 & 61.91 & \textbf{-0.59} & & \textbf{54.65} & \textbf{54.56} & \textbf{-0.09} \\
    \cmidrule{2-10}
    & \multirow{3}*{$L_2$} & AT & 71.57 & 69.99 & -1.58 & & 70.44 & 68.92 & -1.52\\
    & & MLCAT$_{\mathrm{LS}}$ & 75.05 & 74.97 & \textbf{-0.08} & & 55.31 & 55.11 & \textbf{-0.20}\\
    & & MLCAT$_{\mathrm{WP}}$ & \textbf{76.92} & \textbf{76.55} & -0.37 & & \textbf{74.35} & \textbf{73.97} & -0.38 \\
    \bottomrule
  \end{tabular}
  \caption{Test robustness (\%) on SVHN. We omit the standard deviations of 5 runs as they are very small ($< 0.6\%$).}
  \label{table:2}
  \centering
  \begin{tabular}{cclccccccc}
    \toprule
    \multirow{2}*{Network} & \multirow{2}*{Threat Model} & \multirow{2}*{Method} & \multicolumn{3}{c}{PGD-20} & & \multicolumn{3}{c}{AA} \\
    \cmidrule{4-6}
    \cmidrule{8-10}
    & & & Best & Last & Diff & & Best & Last & Diff \\
    \midrule
    \multirow{6}*{PreAct ResNet-18} & \multirow{3}*{$L_\infty$} & AT & 52.88 & 45.29 & -7.59 & & 45.09 & 40.36 & -4.73 \\
     & & MLCAT$_{\mathrm{LS}}$ & \textbf{64.28} & \textbf{62.30} & \textbf{-1.98} & & 34.48 & 32.33 & -2.15\\
     & & MLCAT$_{\mathrm{WP}}$ & 60.34 & 57.79 & -2.55 & & \textbf{51.90} & \textbf{49.76} & \textbf{-2.14} \\
    \cmidrule{2-10}
    & \multirow{3}*{$L_2$} & AT & 66.68 & 64.75 & -1.93 & & 63.55 & 62.14 & -1.41\\
    & & MLCAT$_{\mathrm{LS}}$ & \textbf{75.32} & \textbf{74.06} & -1.26 & & 53.35 & 52.29 & -1.06\\
    & & MLCAT$_{\mathrm{WP}}$ & 72.58 & 71.59 & \textbf{-0.99} & & \textbf{67.21} & \textbf{66.27} & \textbf{-0.94} \\
     \midrule
    \multirow{6}*{Wide ResNet-34-10} & \multirow{3}*{$L_\infty$} & AT & 55.72 & 50.44 & -5.28 & & 48.00 & 42.41 & -5.59\\
    & & MLCAT$_{\mathrm{LS}}$ & \textbf{78.96} & \textbf{77.38} & -1.58 & & 34.66 & 34.21 & -0.45\\
    & & MLCAT$_{\mathrm{WP}}$ & 63.18 & 61.71 & \textbf{-1.47} & & \textbf{54.29} & \textbf{53.93} & \textbf{-0.36} \\
    \cmidrule{2-10}
    & \multirow{3}*{$L_2$} & AT & 67.29 & 65.18 & -2.11 & & 62.88 & 61.06 & -1.82\\
    & & MLCAT$_{\mathrm{LS}}$ & \textbf{85.00} & \textbf{83.47} & -1.53 & & 55.74 & 54.15 & -1.59\\
    & & MLCAT$_{\mathrm{WP}}$ & 75.43 & 74.08 & \textbf{-1.35} & & \textbf{68.91} & \textbf{68.12} & \textbf{-0.79} \\
    \bottomrule
  \end{tabular}
\end{table*}

\begin{table*}[t]
\small
  \caption{Test robustness (\%) on CIFAR100. We omit the standard deviations of 5 runs as they are very small ($< 0.6\%$).}
  \label{table:3}
  \centering
  \begin{tabular}{cclccccccc}
    \toprule
    \multirow{2}*{Network} & \multirow{2}*{Threat Model} & \multirow{2}*{Method} & \multicolumn{3}{c}{PGD-20} & & \multicolumn{3}{c}{AA} \\
    \cmidrule{4-6}
    \cmidrule{8-10}
    & & & Best & Last & Diff & & Best & Last & Diff \\
    \midrule
    \multirow{6}*{PreAct ResNet-18} & \multirow{3}*{$L_\infty$} & AT & 28.01 & 20.39 & -7.62 & & 23.61 & 18.41 & -5.20 \\
     & & MLCAT$_{\mathrm{LS}}$ & 20.09 & 18.14 & -1.95 & & 13.41 & 11.35 & -2.06\\
     & & MLCAT$_{\mathrm{WP}}$ & \textbf{31.27} & \textbf{30.57} & \textbf{-0.70} & & \textbf{25.66} & \textbf{25.28} & \textbf{-0.38} \\
    \cmidrule{2-10}
    & \multirow{3}*{$L_2$} & AT & 41.38 & 35.34 & -6.04 & & 37.94 & 33.58 & -4.36\\
    & & MLCAT$_{\mathrm{LS}}$ & 31.23 & 30.80 & \textbf{-0.43} & & 22.06 & 21.72 & -0.34\\
    & & MLCAT$_{\mathrm{WP}}$  & \textbf{45.49} & \textbf{44.84} & -0.65 & & \textbf{41.22} & \textbf{41.15} & \textbf{-0.07} \\
     \midrule
    \multirow{6}*{Wide ResNet-34-10} & \multirow{3}*{$L_\infty$} & AT & 30.74 & 24.89 & -5.85 & & 26.98 & 23.07 & -3.91\\
    & & MLCAT$_{\mathrm{LS}}$ & 22.86 & 22.18 & -0.68 & & 14.61 & 14.05 & -0.56\\
    & & MLCAT$_{\mathrm{WP}}$ & \textbf{34.97} & \textbf{34.64} & \textbf{-0.33} & & \textbf{29.49} & \textbf{29.25} & \textbf{-0.24} \\
    \cmidrule{2-10}
    & \multirow{3}*{$L_2$} & AT & 44.12 & 41.29 & -2.83 & & 41.39 & 39.34 & -2.05\\
    & & MLCAT$_{\mathrm{LS}}$ & 34.09 & 33.66 & \textbf{-0.43} & & 25.06 & 24.31 & -0.75\\
    & & MLCAT$_{\mathrm{WP}}$ & \textbf{50.17} & \textbf{49.51} & -0.66 & & \textbf{46.05} & \textbf{45.77} & \textbf{-0.28} \\
    \bottomrule
  \end{tabular}
\end{table*}

\begin{figure*}[t]
\centering
    \subfigure[Impact of minimum loss condition $\ell_{min}$]{
        \includegraphics[width=0.48\columnwidth]{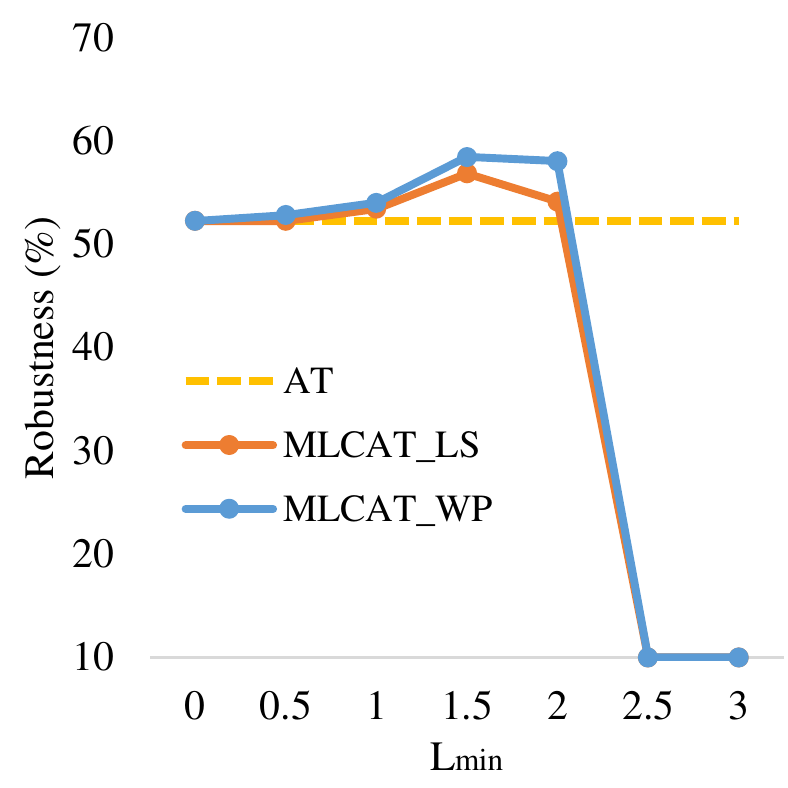}
        \includegraphics[width=0.48\columnwidth]{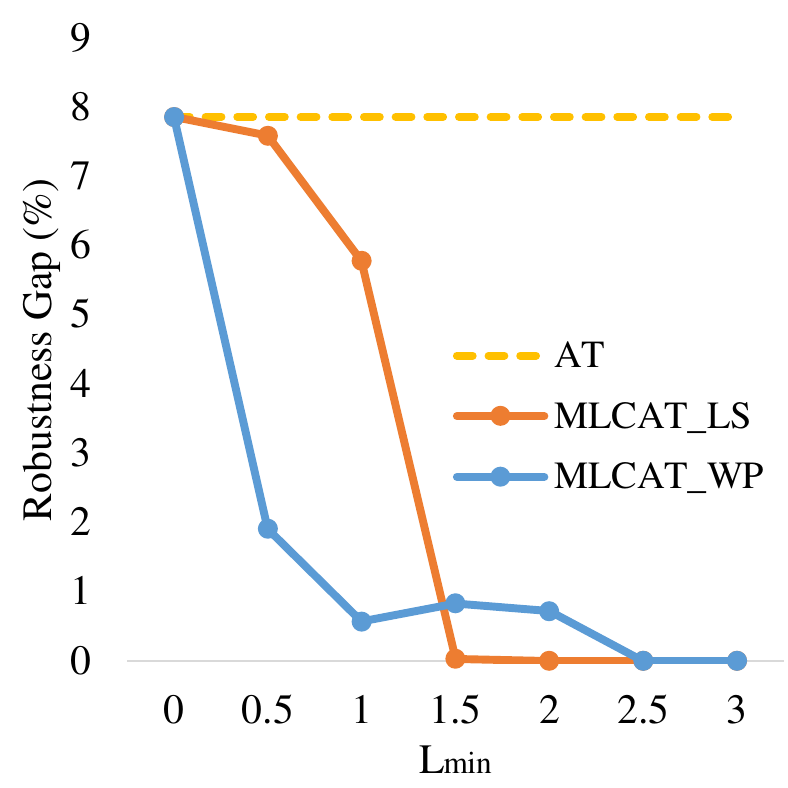}
    }
    \subfigure[Role of loss adjustment strategy $\mathcal{S}$]{
        \includegraphics[width=0.48\columnwidth]{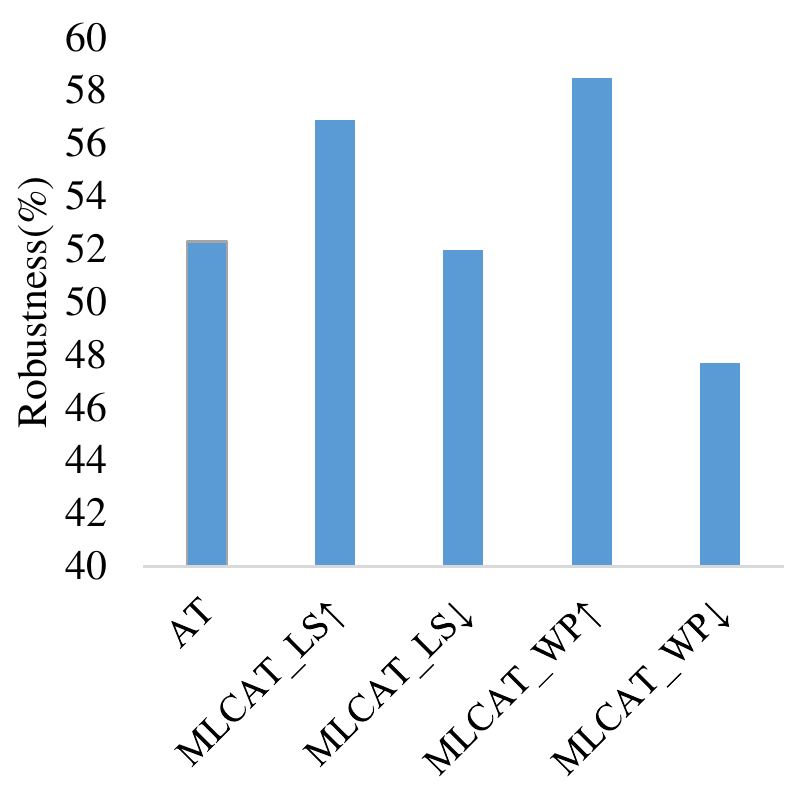}
        \includegraphics[width=0.48\columnwidth]{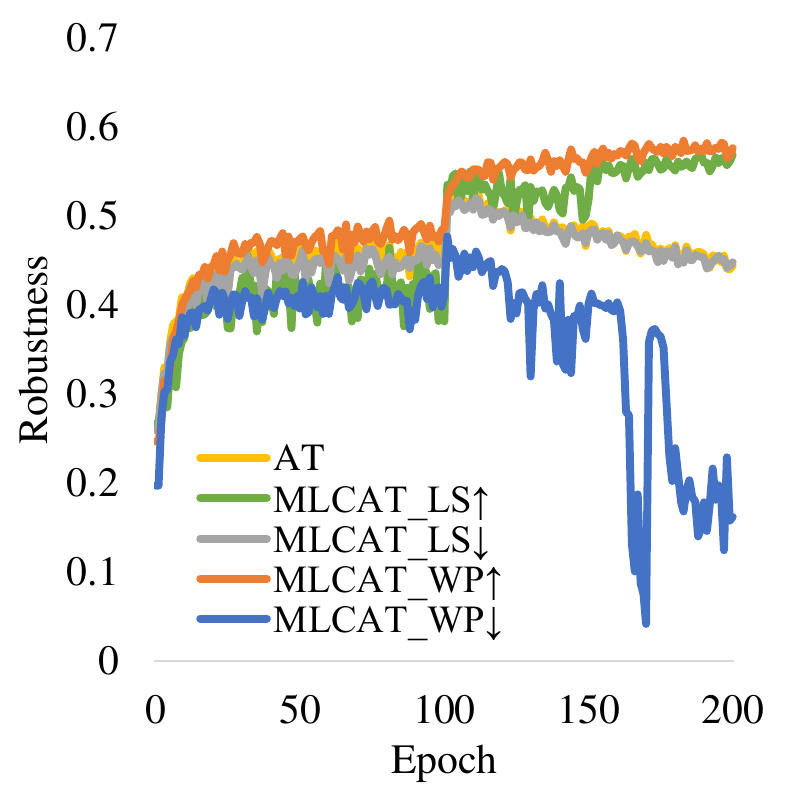}
    }
\caption{The ablation study experiment results on CIFAR10 dataset.}
\label{fig:ablation}
\end{figure*}

\section{Experiment}
\label{Experiment}
In this section, we conduct extensive  experiments to verify the effectiveness of MLCAT$_{\mathrm{LS}}$ and MLCAT$_{\mathrm{WP}}$ including their experimental settings (Section \ref{experiment setting}), performance evaluation (Section \ref{Performance Evaluation}) and ablation studies (Section \ref{ablation}).

\subsection{Experimental Settings}
\label{experiment setting}
Our implementation is based on PyTorch and the code is publicly available\footnote{\url{https://github.com/ChaojianYu/Understanding-Robust-Overfitting}}. We conduct extensive experiments across three benchmark datasets (CIFAR10 \cite{krizhevsky2009learning}, SVHN \cite{netzer2011reading} and CIFAR100 \cite{krizhevsky2009learning}) and two threat models ($L_\infty$ and $L_2$). We use PreAct ResNet-18 \cite{he2016deep} and Wide ResNet-34-10 \cite{zagoruyko2016wide} as the network structure following \cite{rice2020overfitting}. For training, the model is trained for 200 epochs using SGD with momentum 0.9, weight decay $5 \times 10^{-4}$, and an initial learning rate of 0.1. The learning rate is divided by 10 at the 100-th and 150-th epoch, respectively. Standard data augmentation including random crops with 4 pixels of padding and random horizontal flips are applied for CIFAR10 and CIFAR100, and no data augmentation is used on SVHN. For adversary, 10-step PGD attack is applied: for $L_\infty$ threat model, perturbation size $\epsilon=8/255$, step size $\alpha=1/255$ for SVHN, and $\alpha=2/255$ for both CIFAR10 and CIFAR100; for $L_2$ threat model, perturbation size $\epsilon=128/255$, step size $\alpha=15/255$ for all datasets, which is a standard setting for PGD-based adversarial training \cite{madry2017towards}. For testing, model robustness is evaluated by measuring the accuracy on test data under different adversarial attacks, including 20-step PGD (PGD-20) \cite{madry2017towards} and Auto Attack (AA) \cite{croce2020reliable}. AA is regarded as the most reliable robustness evaluation to date, which is an ensemble of complementary attacks, consisting of three white-box attacks (APGD-CE \citep{croce2020reliable}, APGD-DLR \citep{croce2020reliable}, and FAB \citep{croce2020minimally}) and a black-box attack (Square Attack \citep{andriushchenko2020square}). The degree of robust overfitting is evaluated by the robust accuracy gap during training. For hyperparameter in MLCAT$_{\mathrm{LS}}$ and MLCAT$_{\mathrm{WP}}$, we set the minimum loss conditions $\ell_{min}=1.5$ for CIFAR10 and SVHN, and $\ell_{min}=4.0$ for CIFAR100. Other hyperparameters of the baselines are configured as per their original papers.

\subsection{Performance Evaluation}\label{Performance Evaluation}

In this part, we report the experimental results of MLCAT$_{\mathrm{LS}}$ and MLCAT$_{\mathrm{WP}}$, and more experimental results are provided in Appendix \ref{AP_C1}.

\textbf{CIFAR10 Results.} The evaluation results on CIFAR10 dataset are summarized in Table \ref{table:1}, where ``Best'' is the highest robustness that ever achieved during training; ``Last'' is the test robustness at the last epoch checkpoint; ``Diff'' denotes the robust accuracy gap between the ``Best'' and ``Last''. First, it is observed that both MLCAT$_{\mathrm{LS}}$ and MLCAT$_{\mathrm{WP}}$ achieve superior robustness performance under PGD-20 attack. Then, for AA attack, MLCAT$_{\mathrm{WP}}$ can still boost adversarial robustness, while MLCAT$_{\mathrm{LS}}$ achieves the worse robustness performance than AT. This is because loss scaling technique make network sensitive to the logit scaling attack \cite{hitaj2021evaluating}. Finally and most importantly, MLCAT$_{\mathrm{LS}}$ and MLCAT$_{\mathrm{WP}}$ significantly narrow the robustness gaps under both PGD-20 attack and AA attack, which indicates they can effectively eliminate robust overfitting in adversarial training across different network architectures and threat models.

\textbf{SVHN Results.} We further report results on the SVHN dataset, which are summarized in Table \ref{table:2}. Experimental results show that the proposed method improve adversarial robustness and narrow the robustness gap by a large margin under both PGD-20 attack and AA attack, demonstrating the effectiveness of the proposed MLCAT prototype.

\textbf{CIFAR100 Results.} We also conduct experiments on CIFAR100 dataset. Note that this dataset is more challenging than CIFAR10 as the number of classes/training images per class is ten times larger/smaller than that of CIFAR10. As shown by the results in Table \ref{table:3}, the proposed methods are still able to eliminate robust overfitting and improve adversarial robustness even on more difficult datasets. It verifies that MLCAT prototype eliminates robust overfitting reliably and is general across different datasets, network architectures and threat models.

\subsection{Ablation Studies}\label{ablation}
In this part, we investigate the impacts of algorithmic components using PreAct ResNet-18 on CIFAR10 under $L_\infty$ threat model following the same experimental setting as Section \ref{experiment setting}.

\textbf{The Impact of Minimum Loss Condition $\ell_{min}$.} To validate the effectiveness of introducing minimum loss constraint in our MLCAT prototype, we investigate the effect of different $\ell_{min}$ for the robustness performance and robustness gap (the gap between ``best'' and ``last'' robust accuracy). The value of minimum loss condition $\ell_{min}$ vary from 0 to 3.0, and the results are summarized in Figure \ref{fig:ablation}(a). As expected, increasing $\ell_{min}$ leads to the smaller robustness gap. For robustness performance, when $\ell_{min}$ is small, increasing $\ell_{min}$ leads to the higher robust accuracy than AT. When $\ell_{min}$ is greater than 1.5, it is observed increasing $\ell_{min}$ makes model robustness decrease and even leads to the training collapses, implying that the additional measures, such as loss scaling and weight perturbation, are inherently detrimental to robustness improvement. It sheds light on the importance of $\ell_{min}$, whose responsibility is to distinguish between small-loss data and large-loss data in 
the MLCAT prototype. Similar pattern can also be observed in SVHN and CIFAR100 dataset (shown in Appendix \ref{AP_C2}). Therefore, we uniformly adopt $\ell_{min}=1.5$ for CIFAR10 and SVHN, and $\ell_{min}=4.0$ for CIFAR100 in consideration of the elimination of robust overfitting as well as robustness improvement.

\textbf{The Role of Loss Adjustment Strategy $\mathcal{S}$.} We further investigate the role of loss adjustment strategy within the MLCAT prototype by comparing several different schemes:
1) additive mapping, which increases the loss of small-loss data by Eq.(\ref{1-increase}) and Eq.(\ref{2-increase}). We denote them as MLCAT$_{\mathrm{LS\uparrow}}$ and MLCAT$_{\mathrm{WP\uparrow}}$, respectively;
2) identical mapping, which keeps the loss of small-loss data unchanged (equivalent to standard AT);
3) subtractive mapping, which decreases the loss of small-loss data. They are implemented by dividing the scaling coefficient $\frac{\ell_{min}}{\ell_i}$ in Eq.(\ref{1-increase}) and subtracting the weight perturbation $v$ in Eq.(\ref{2-increase}), which are denoted as MLCAT$_{\mathrm{LS\downarrow}}$ and MLCAT$_{\mathrm{WP\downarrow}}$, respectively. 
Their robustness performance and test accuracy curves are summarized in Figure \ref{fig:ablation}(b). It is observed that decreasing the loss of small-loss data not only fails to suppress robust overfitting but also leads to worse adversarial robustness. In contrast, increasing the loss of small-loss data not only eliminate robust overfitting but also facilitates models to learn these data and further improve adversarial robustness. These comparisons echo our approach's philosophy of turning waste into treasure and making full use of each adversarial data.

\section{Conclusion}
\label{Conclusion}
In this paper, we investigate robust overfitting from the perspective of data distribution and identify that some small-loss data lead to robust overfitting under strong adversary modes. Following this, we propose \emph{minimum loss constrained adversarial training} (MLCAT) prototype. The proposed prototype distinguish itself from others by using additional measures to increase the loss of small-loss data, which prevents the model from fitting these data, and thus effectively avoid robust overfitting. We further provide two specific MLCAT implementations: loss scaling derived from loss correction and weight perturbation derived from parameter correction. Comprehensive experiments show that two realizations of MLCAT can eliminate robust overfitting and improve adversarial robustness across different network architectures, threat models and benchmark datasets.

\section*{Acknowledgement}
BH is supported by the RGC Early Career Scheme No. 22200720, NSFC Young Scientists Fund No. 62006202, and Guangdong Basic and Applied Basic Research Foundation No. 2022A1515011652.
LS is supported by Science and Technology Innovation 2030 –“Brain Science and Brain-like Research” Major Project (No. 2021ZD0201402 and No. 2021ZD0201405).
JY is sponsored by CAAI-Huawei MindSpore Open Fund (CAAIXSJLJJ-2021-016B), Anhui Province Key Research and Development Program (202104a05020007).
CG is supported by NSF of China (No: 61973162), NSF of Jiangsu Province (No: BZ2021013), and the Fundamental Research Funds for the Central Universities (Nos: 30920032202, 30921013114).
MMG is supported by ARC DE210101624. 
TLL is partially supported by Australian Research Council Projects DE-190101473, IC-190100031, and DP-220102121.


\bibliography{example_paper}
\bibliographystyle{icml2022}

\newpage
\appendix
\onecolumn
\section{More Evidences for Robust Overfitting and Data Distribution}
\label{AP_A}
\begin{figure}[h]
\centering
    \subfigure[]{
        \includegraphics[width=0.24\columnwidth]{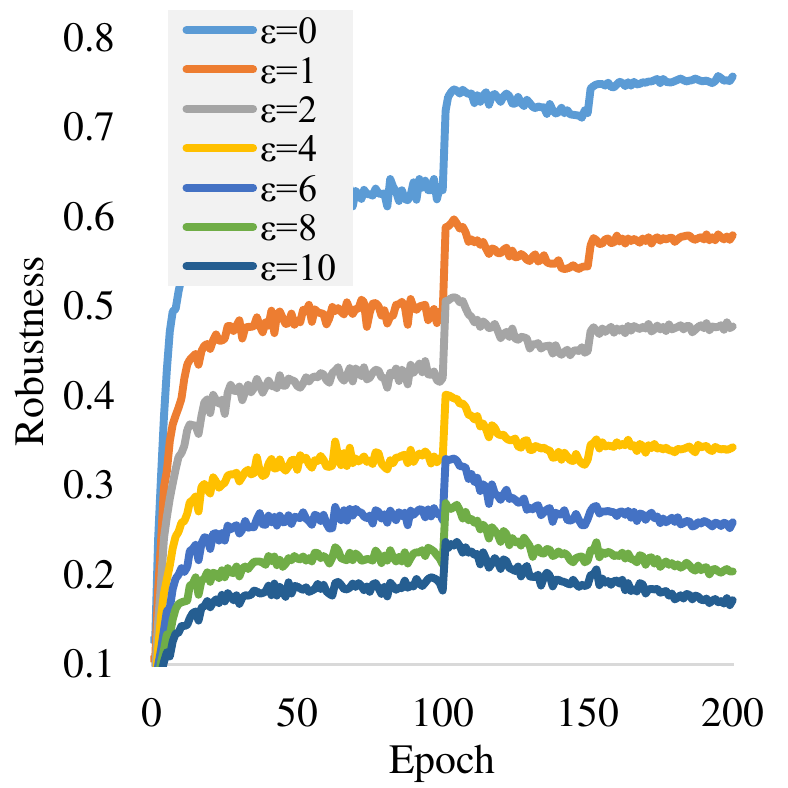}
    }
    \subfigure[Data distribution under perturbation size of 0, 1, and 2 (from left to right)]{
        \includegraphics[width=0.24\columnwidth]{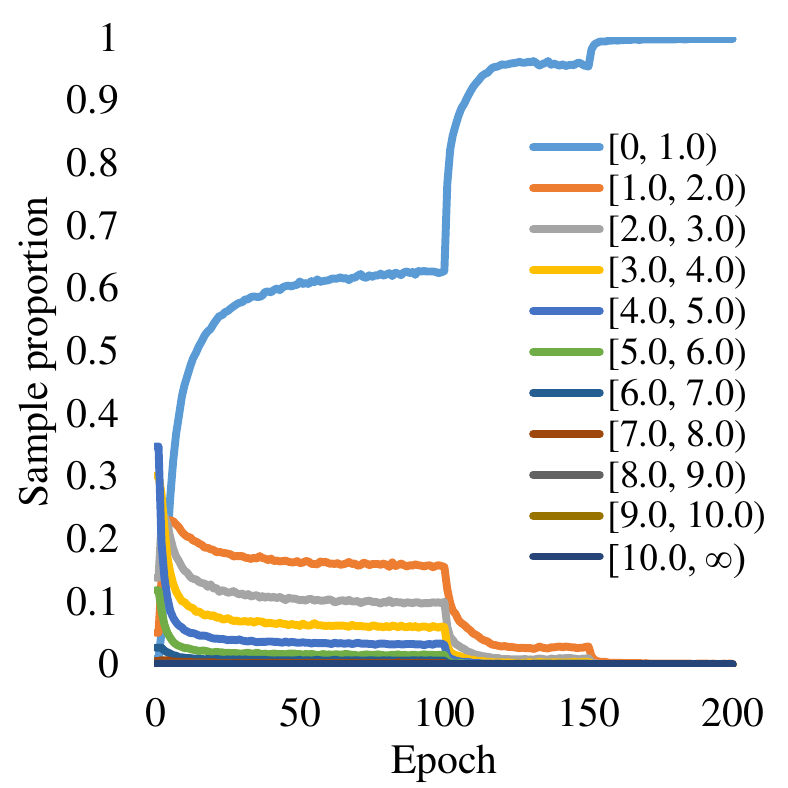}
        \includegraphics[width=0.24\columnwidth]{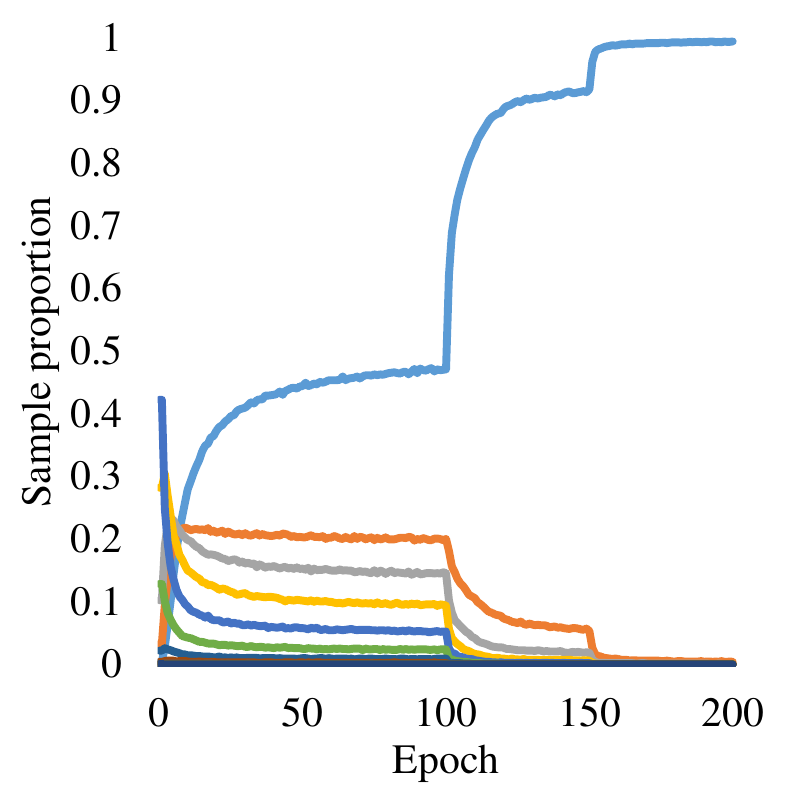}
        \includegraphics[width=0.24\columnwidth]{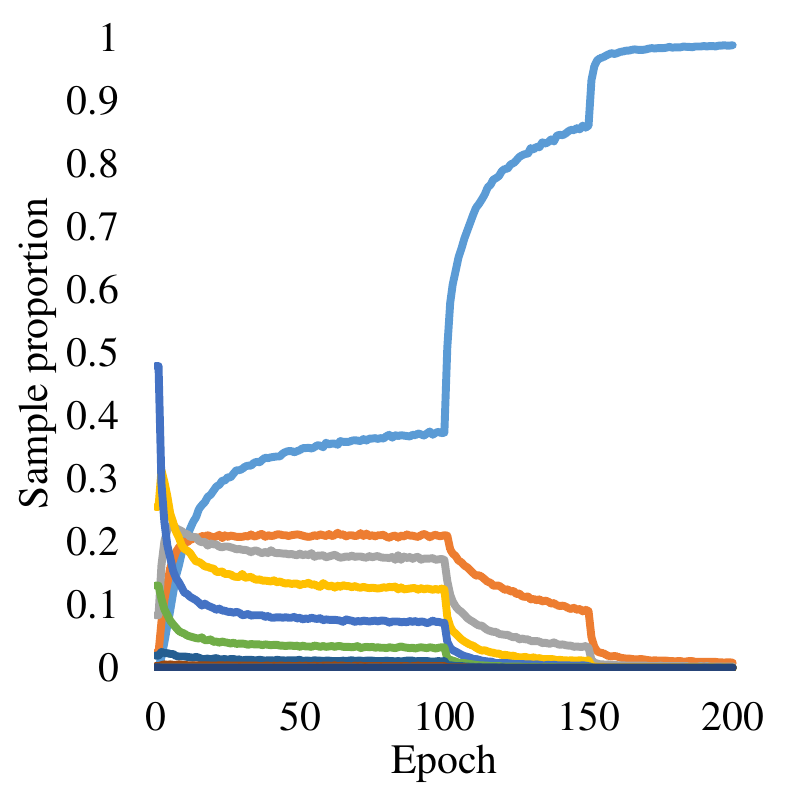}
    }
    \\
    \subfigure[Data distribution under perturbation size of 4, 6, 8, and 10 (from left to right)]{
        \includegraphics[width=0.24\columnwidth]{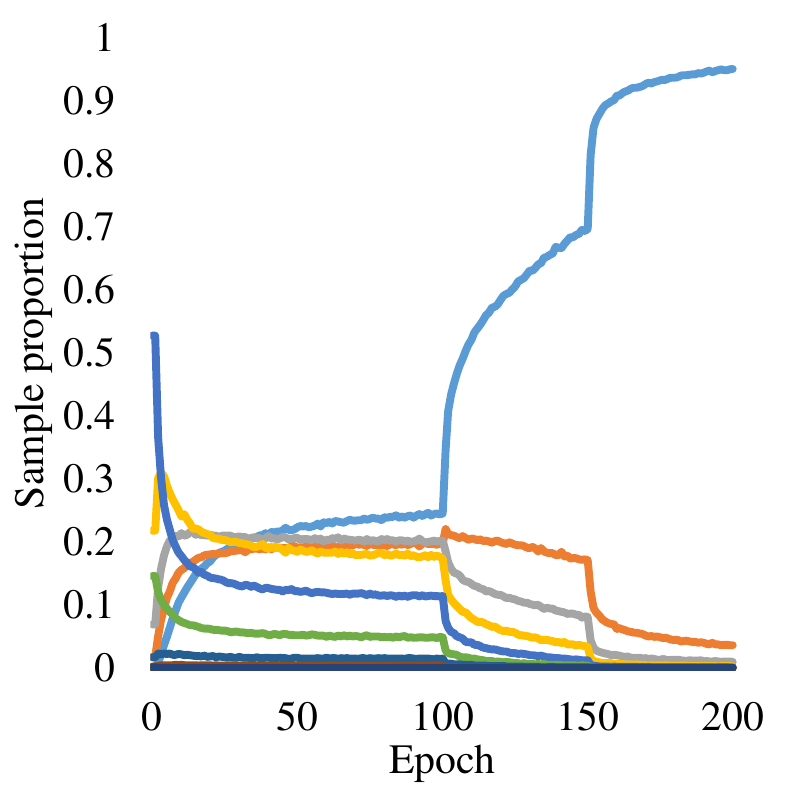}
        \includegraphics[width=0.24\columnwidth]{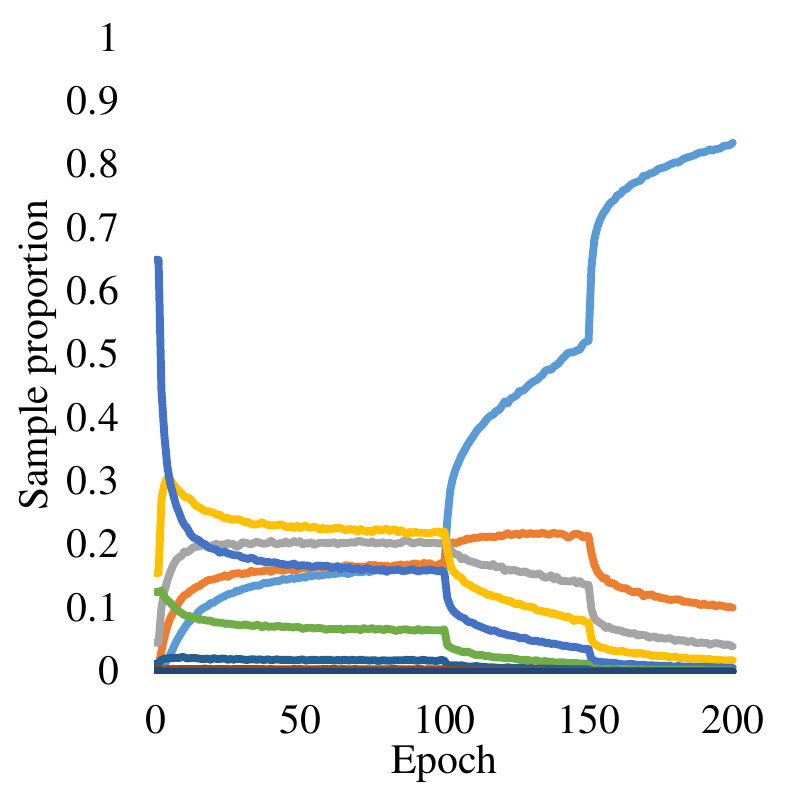}
        \includegraphics[width=0.24\columnwidth]{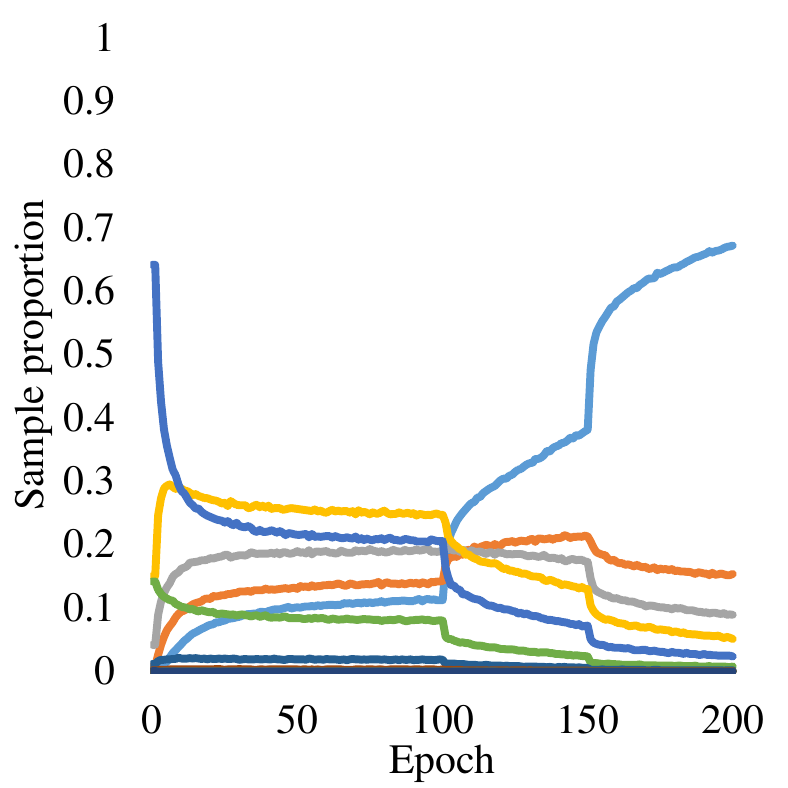}
        \includegraphics[width=0.24\columnwidth]{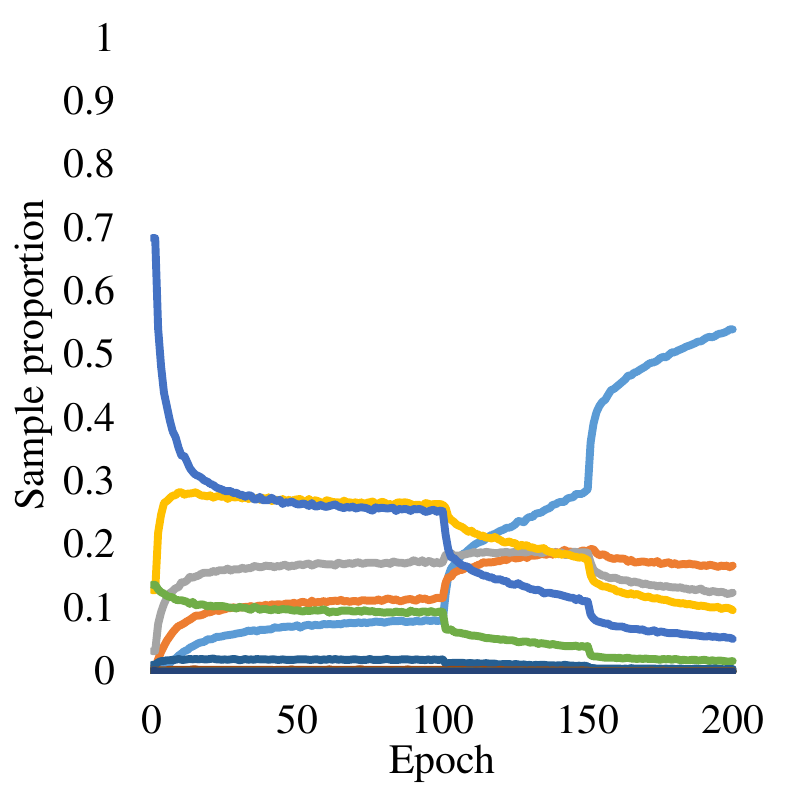}
    }
\caption{Robust overfitting behaviors and data distribution on CIFAR100 using PreAct ResNet-18 under $L_\infty$ threat model. (a): The test robustness of adversarial training with various perturbation size $\epsilon$; (b) and (c): The distribution of training data in different loss ranges under various perturbation size $\epsilon$.}
\label{fig:4}
\end{figure}

\begin{figure}[h]
\centering
    \subfigure[]{
        \includegraphics[width=0.24\columnwidth]{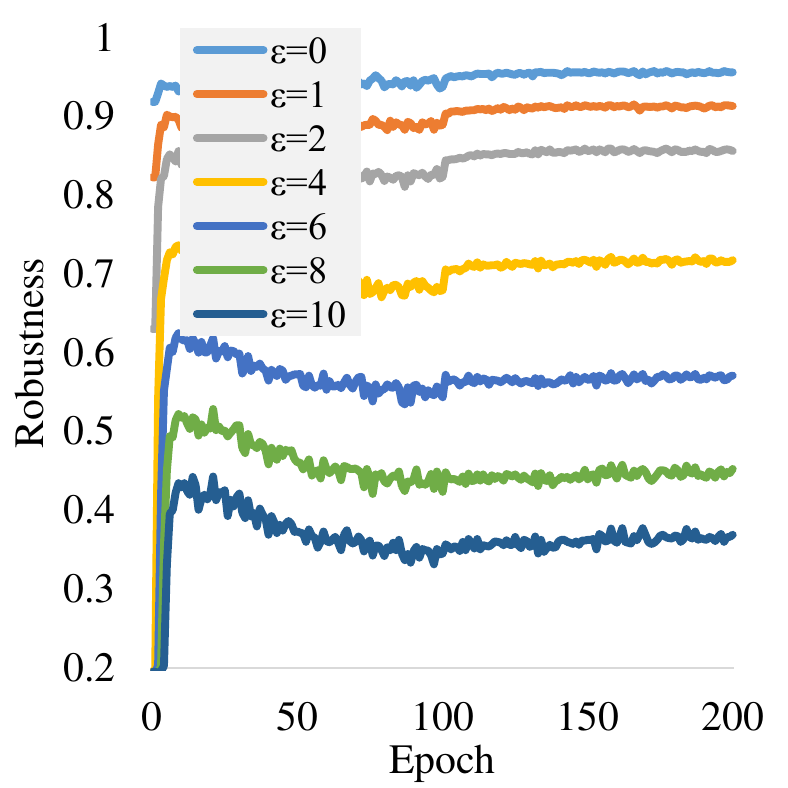}
    }
    \subfigure[Data distribution under perturbation size of 0, 1, and 2 (from left to right)]{
        \includegraphics[width=0.24\columnwidth]{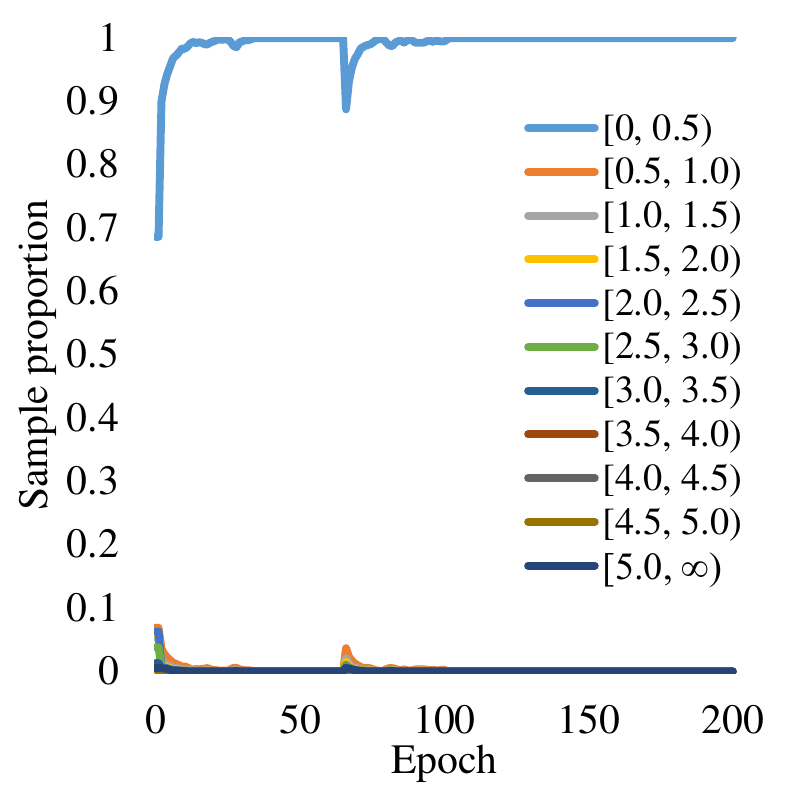}
        \includegraphics[width=0.24\columnwidth]{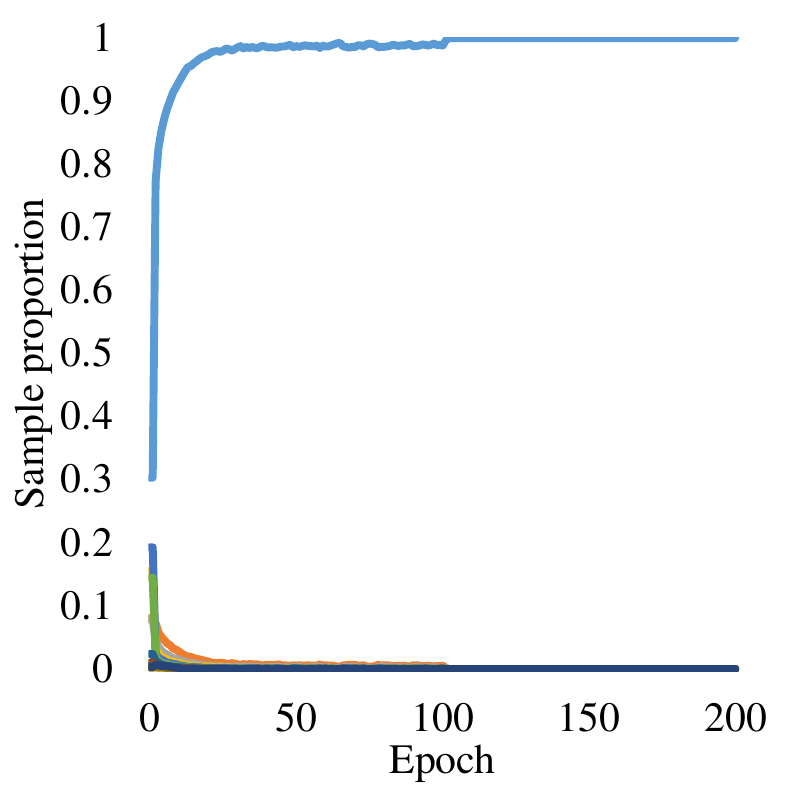}
        \includegraphics[width=0.24\columnwidth]{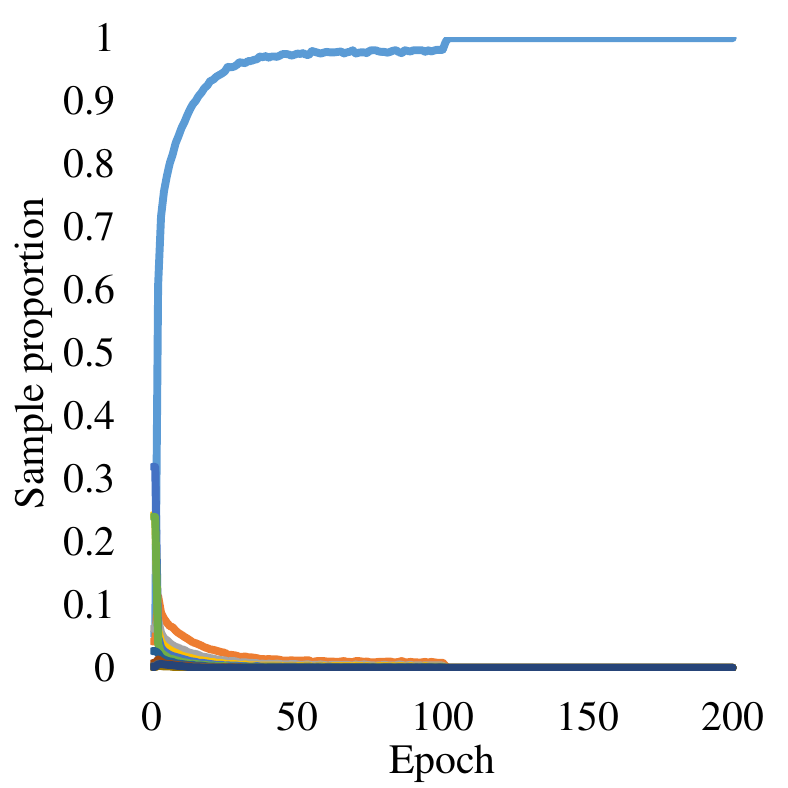}
    }
    \\
    \subfigure[Data distribution under perturbation size of 4, 6, 8, and 10 (from left to right)]{
        \includegraphics[width=0.24\columnwidth]{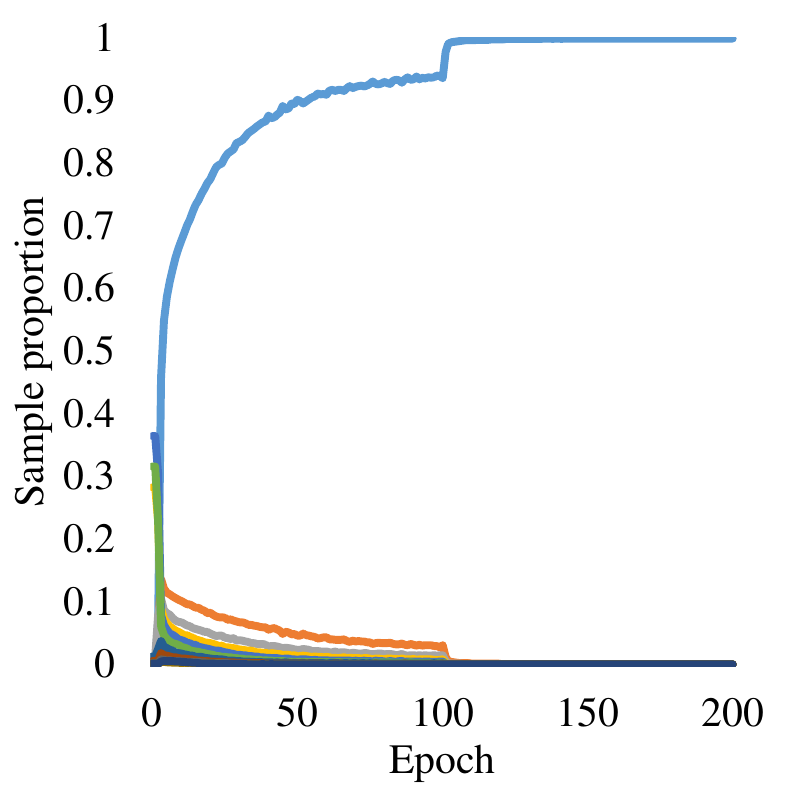}
        \includegraphics[width=0.24\columnwidth]{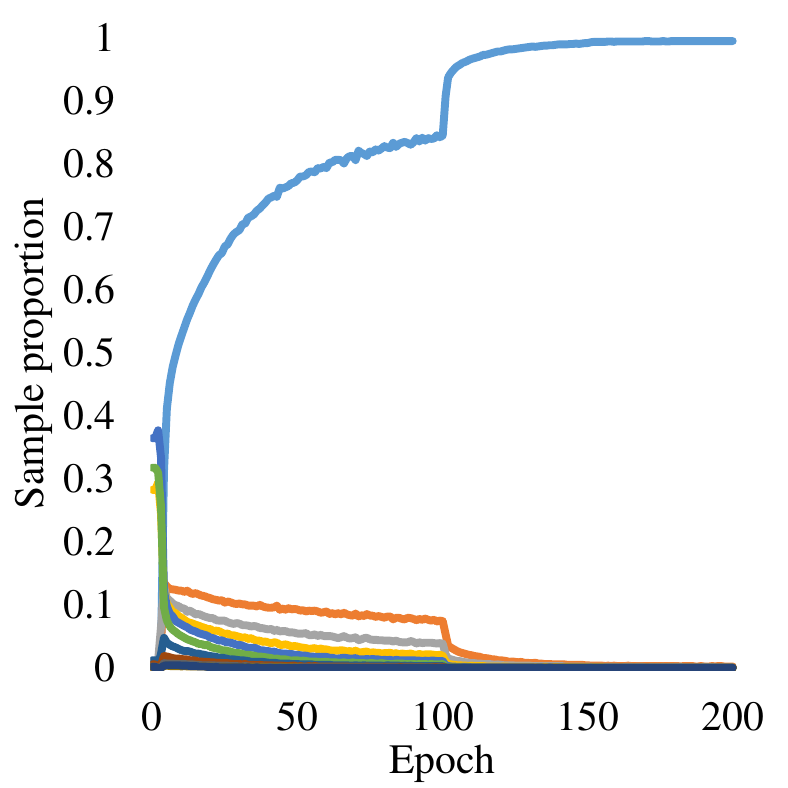}
        \includegraphics[width=0.24\columnwidth]{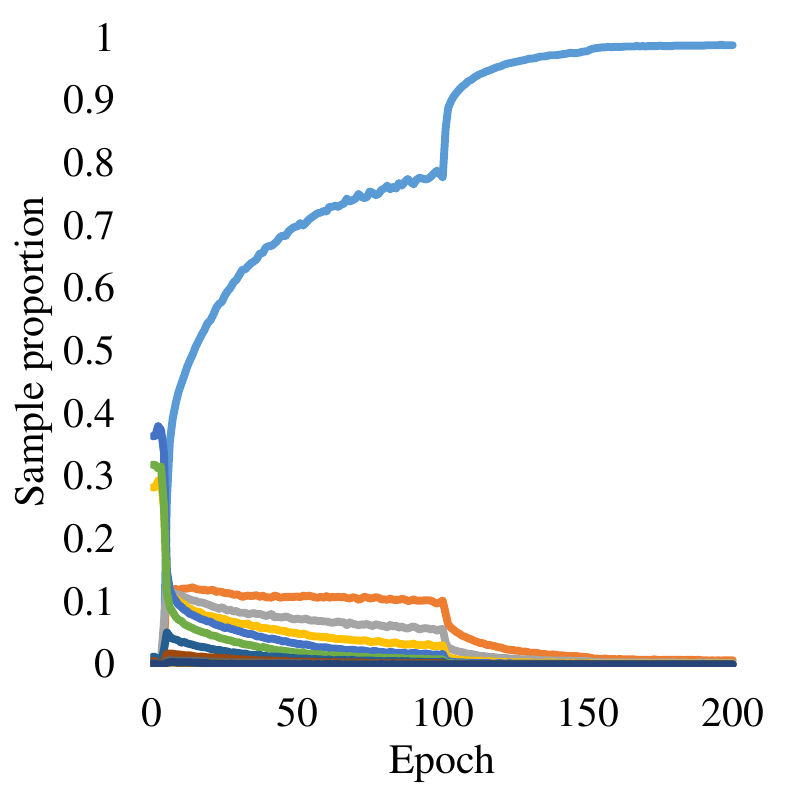}
        \includegraphics[width=0.24\columnwidth]{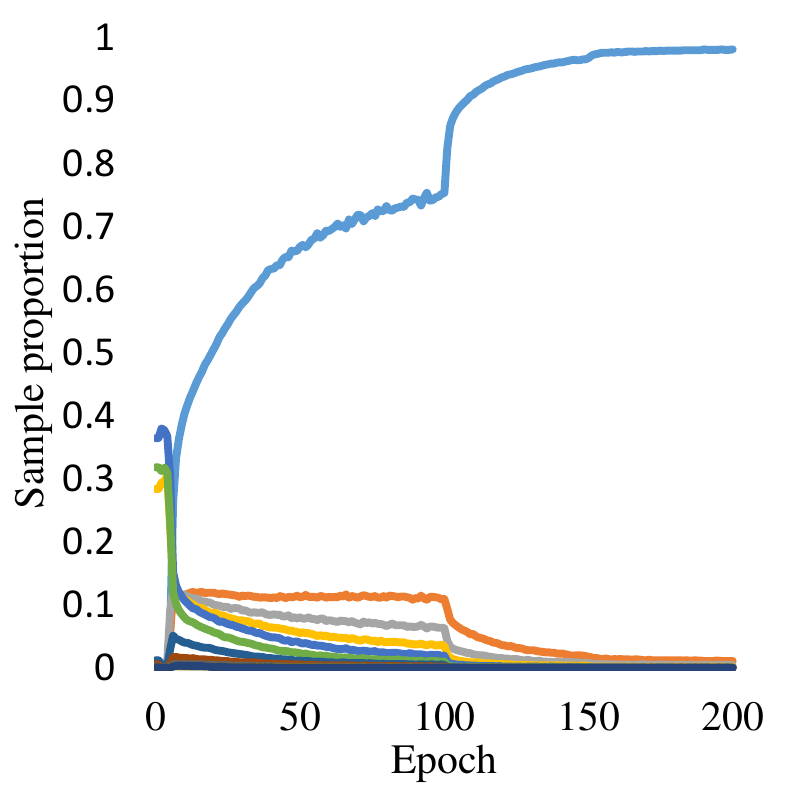}
    }
\caption{Robust overfitting behaviors and data distribution on SVHN using PreAct ResNet-18 under $L_\infty$ threat model. (a): The test robustness of adversarial training with various perturbation size $\epsilon$; (b) and (c): The distribution of training data in different loss ranges under various perturbation size $\epsilon$.}
\label{fig:5}
\end{figure}

\begin{figure}[h]
\centering
    \subfigure[]{
        \includegraphics[width=0.24\columnwidth]{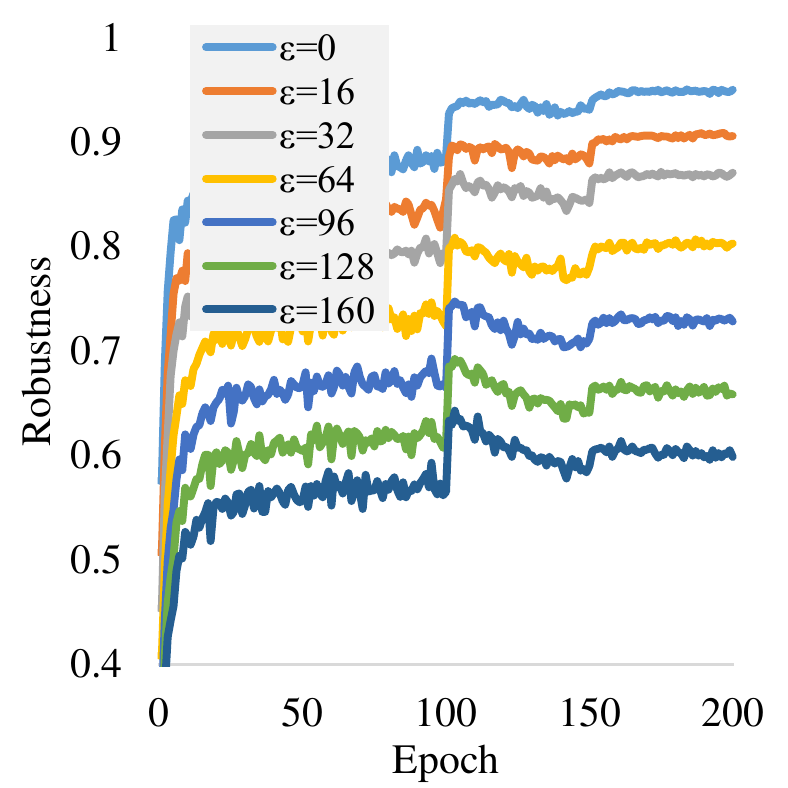}
    }
    \subfigure[Data distribution under perturbation size of 0, 16, and 32 (from left to right)]{
        \includegraphics[width=0.24\columnwidth]{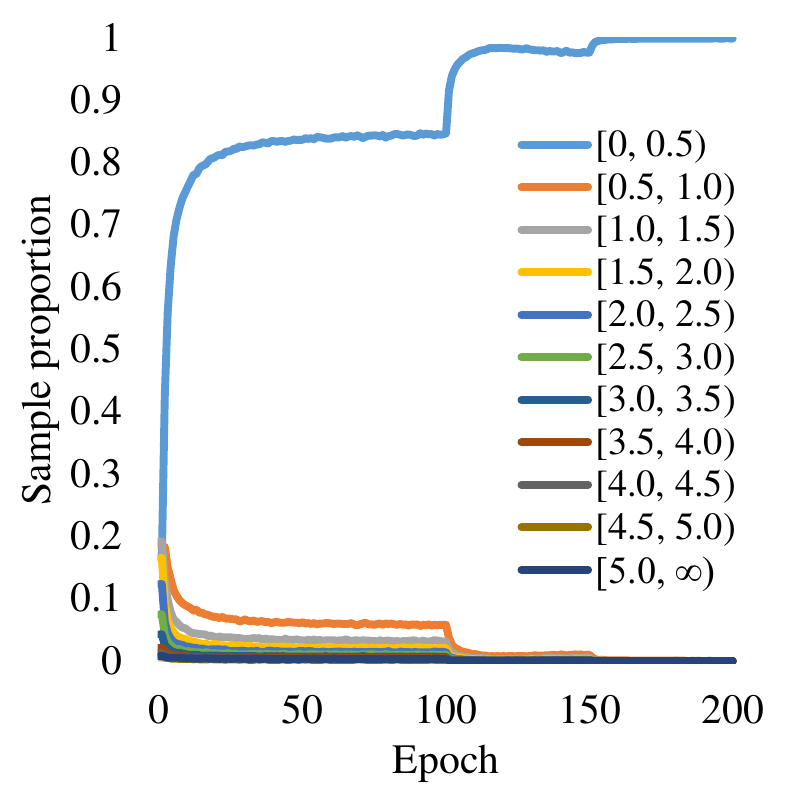}
        \includegraphics[width=0.24\columnwidth]{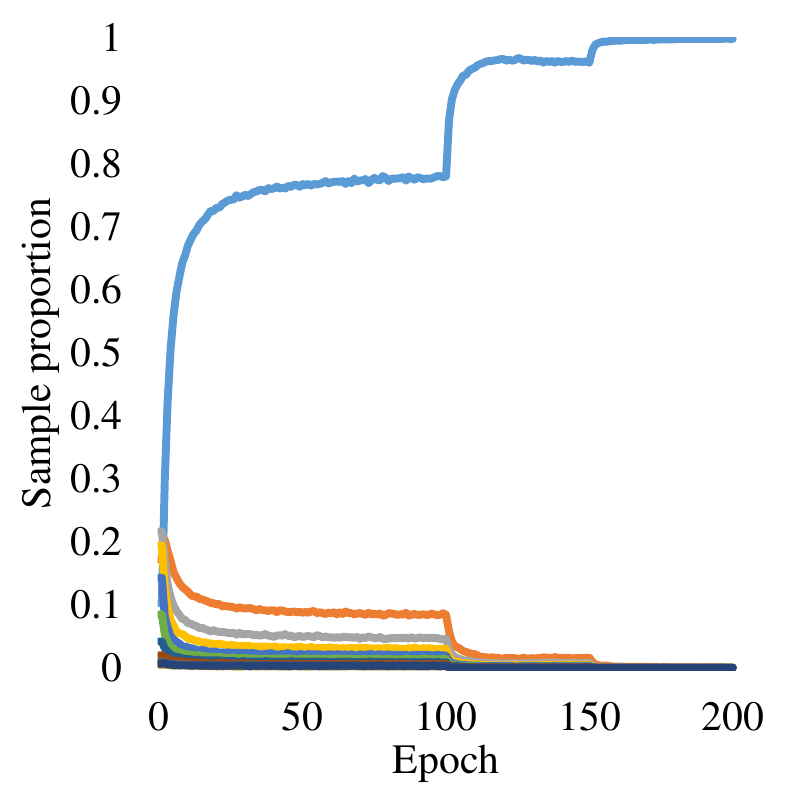}
        \includegraphics[width=0.24\columnwidth]{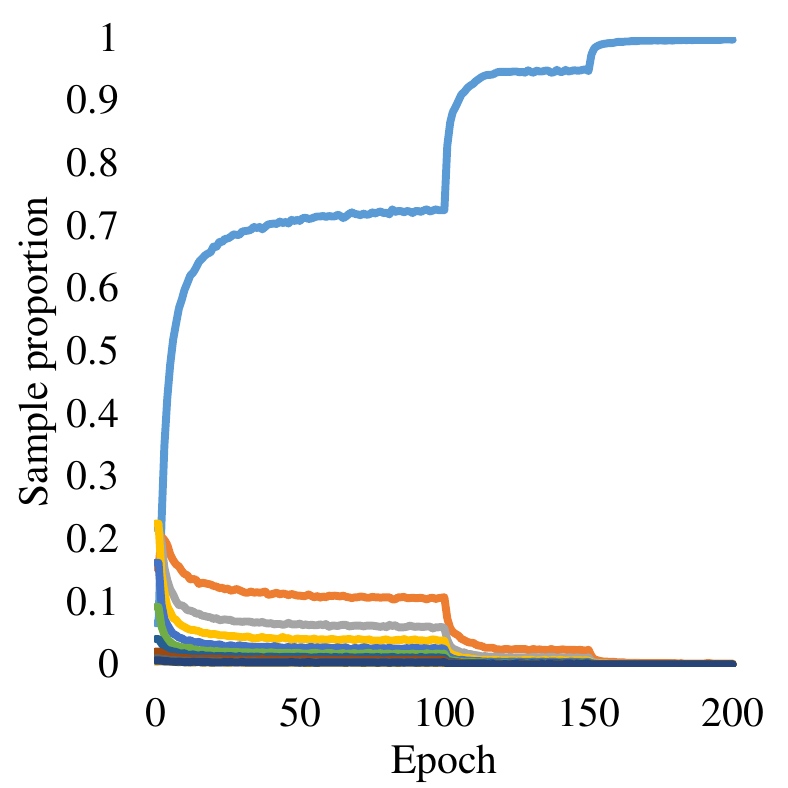}
    }
    \\
    \subfigure[Data distribution under perturbation size of 64, 96, 128, and 160 (from left to right)]{
        \includegraphics[width=0.24\columnwidth]{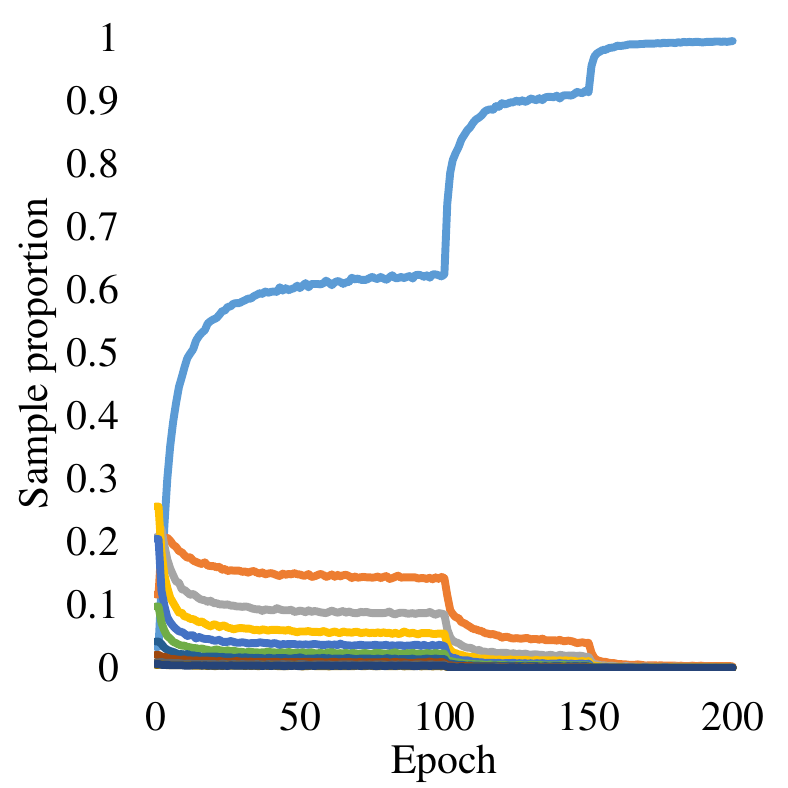}
        \includegraphics[width=0.24\columnwidth]{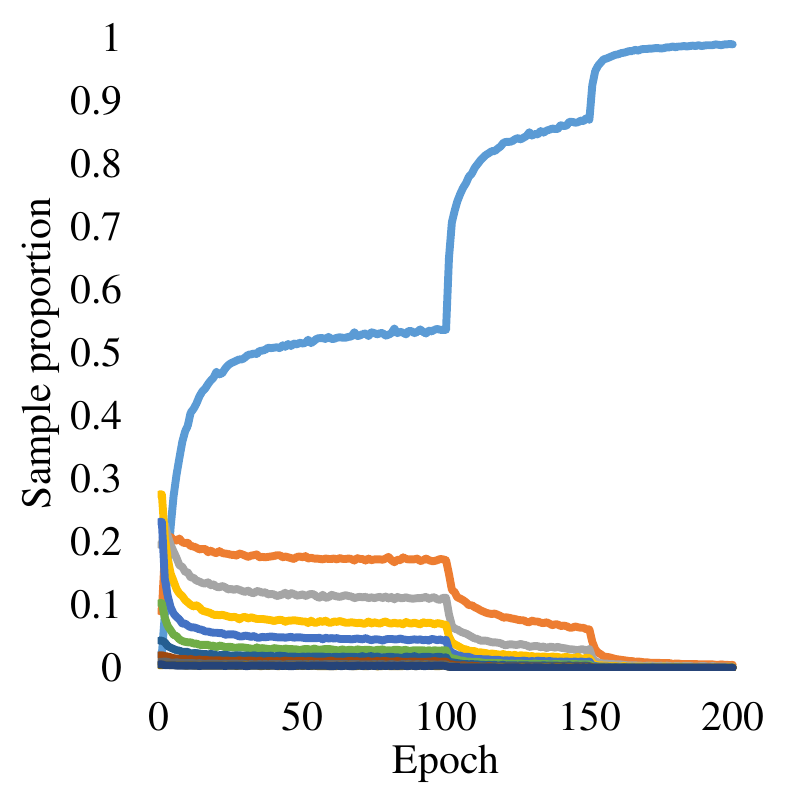}
        \includegraphics[width=0.24\columnwidth]{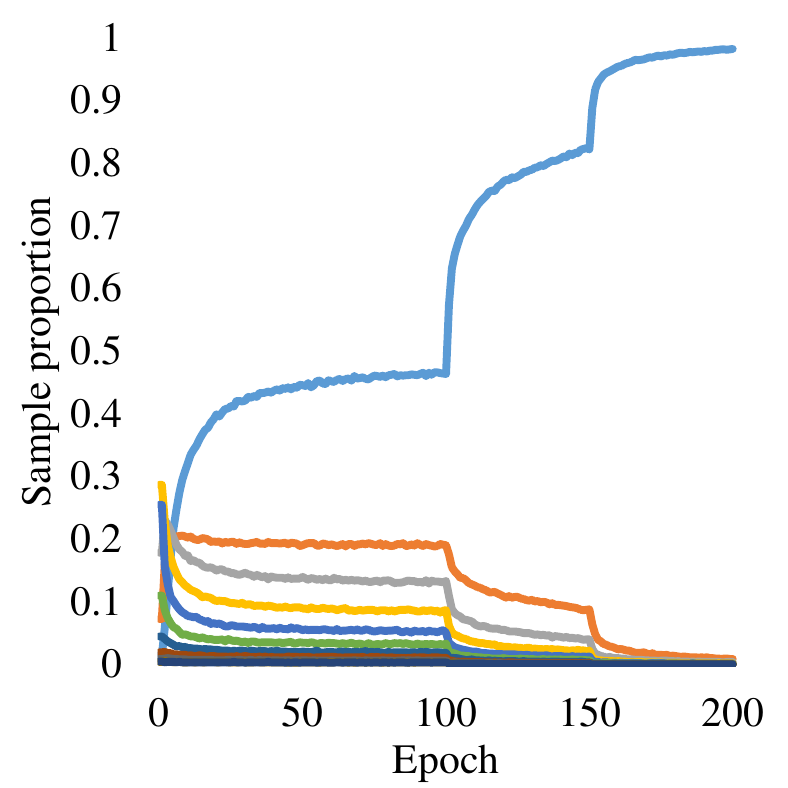}
        \includegraphics[width=0.24\columnwidth]{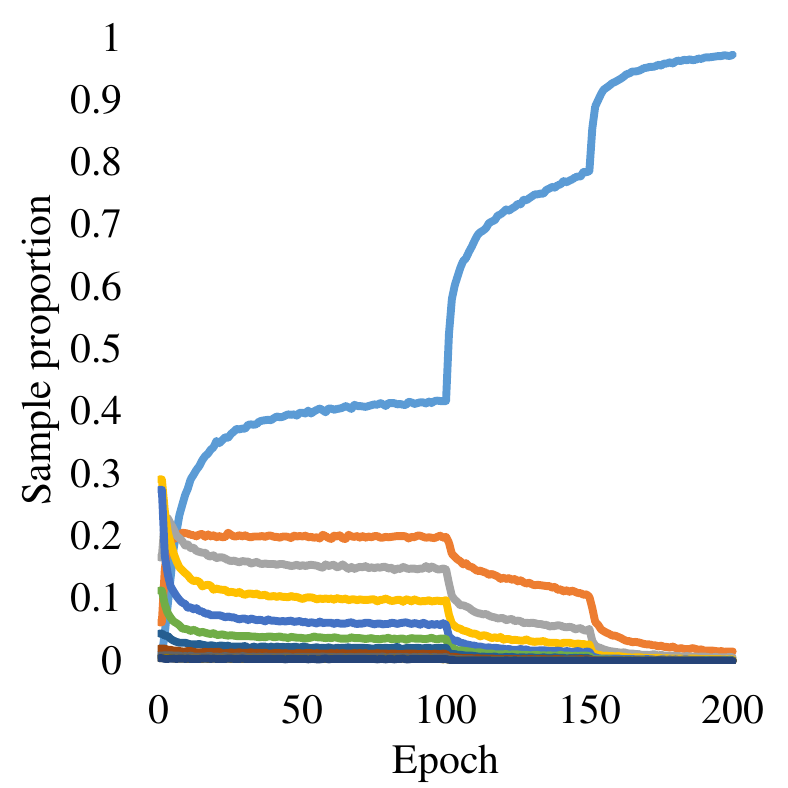}
    }
\caption{Robust overfitting behaviors and data distribution on CIFAR10 using PreAct ResNet-18 under $L_2$ threat model. (a): The test robustness of adversarial training with various perturbation size $\epsilon$; (b) and (c): The distribution of training data in different loss ranges under various perturbation size $\epsilon$.}
\label{fig:6}
\end{figure}

\begin{figure}[h]
\centering
    \subfigure[]{
        \includegraphics[width=0.24\columnwidth]{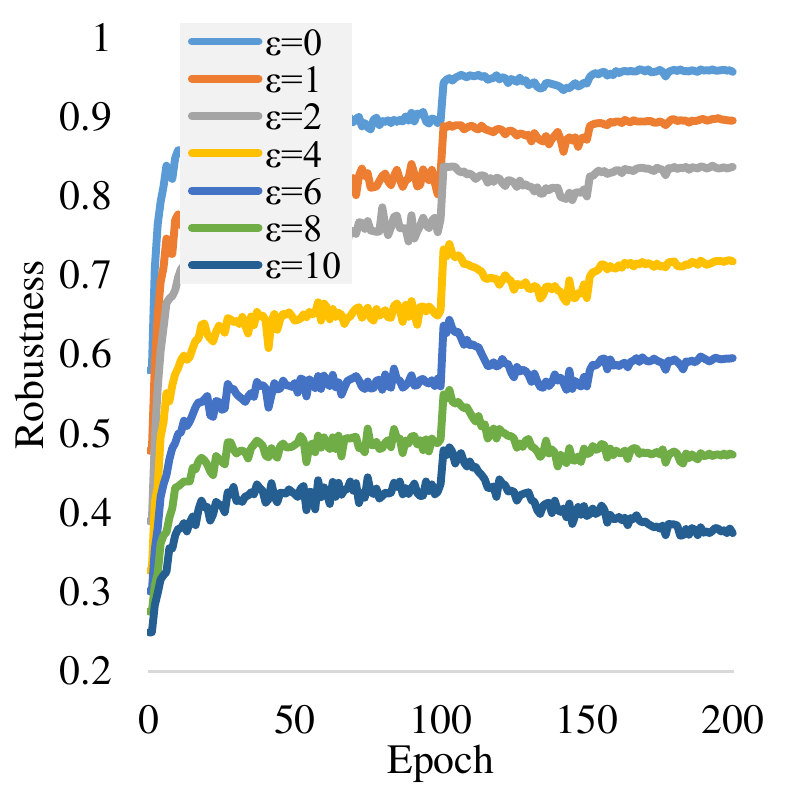}
    }
    \subfigure[Data distribution under perturbation size of 0, 1, and 2 (from left to right)]{
        \includegraphics[width=0.24\columnwidth]{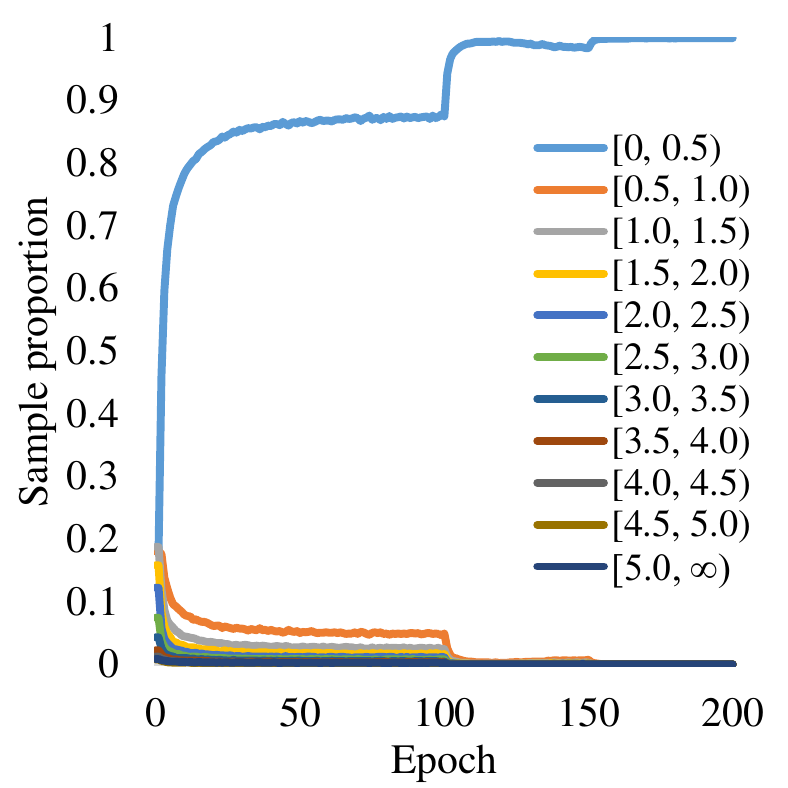}
        \includegraphics[width=0.24\columnwidth]{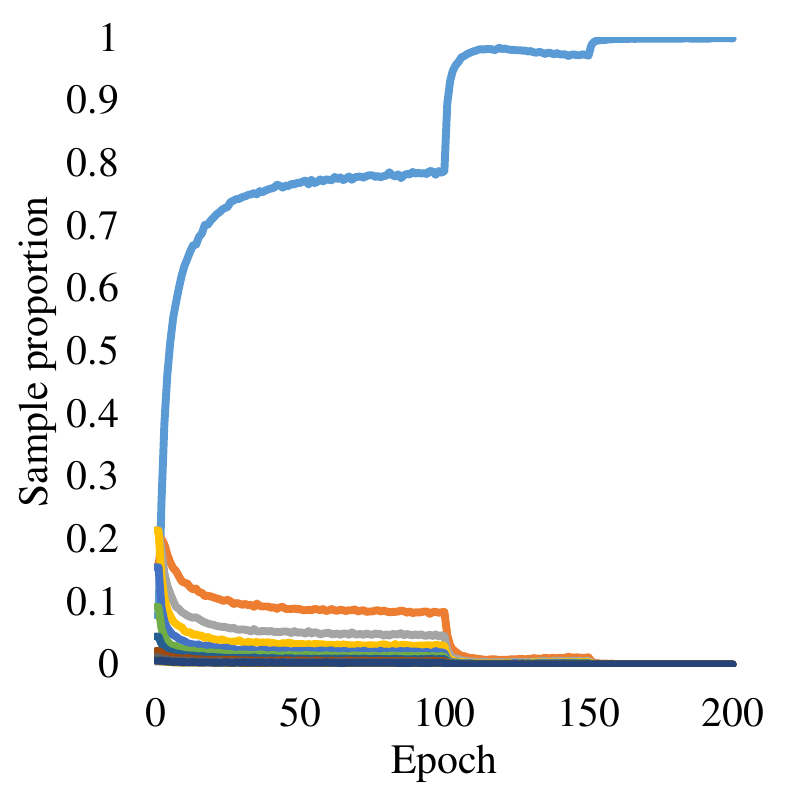}
        \includegraphics[width=0.24\columnwidth]{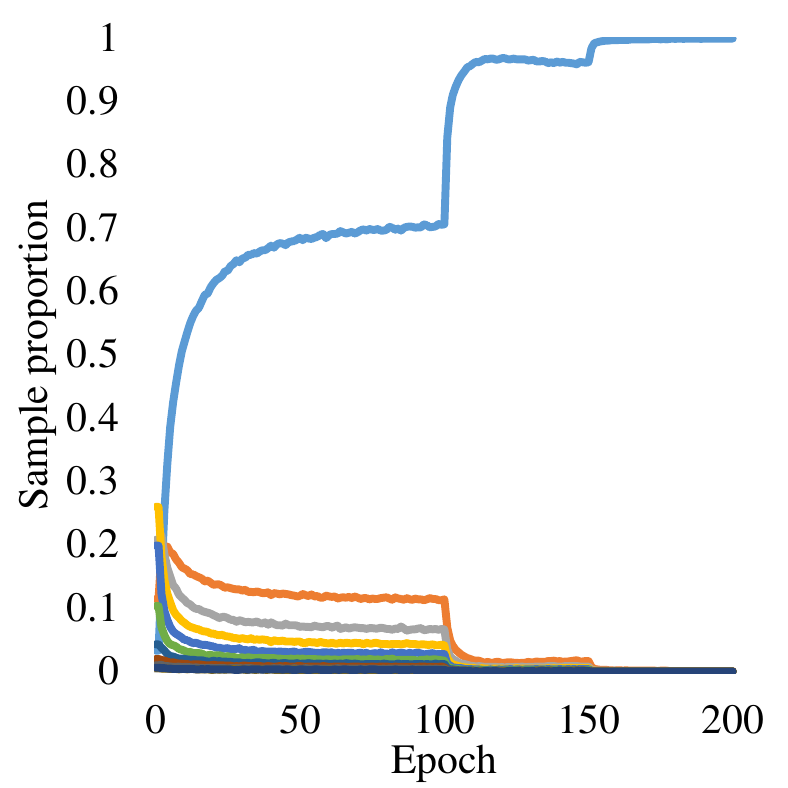}
    }
    \\
    \subfigure[Data distribution under perturbation size of 4, 6, 8, and 10 (from left to right)]{
        \includegraphics[width=0.24\columnwidth]{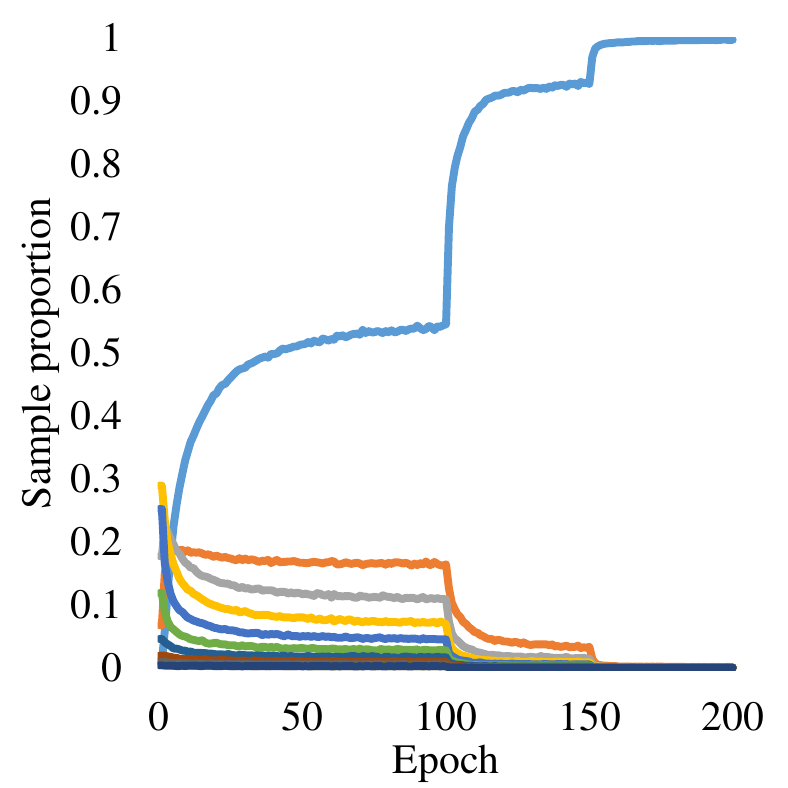}
        \includegraphics[width=0.24\columnwidth]{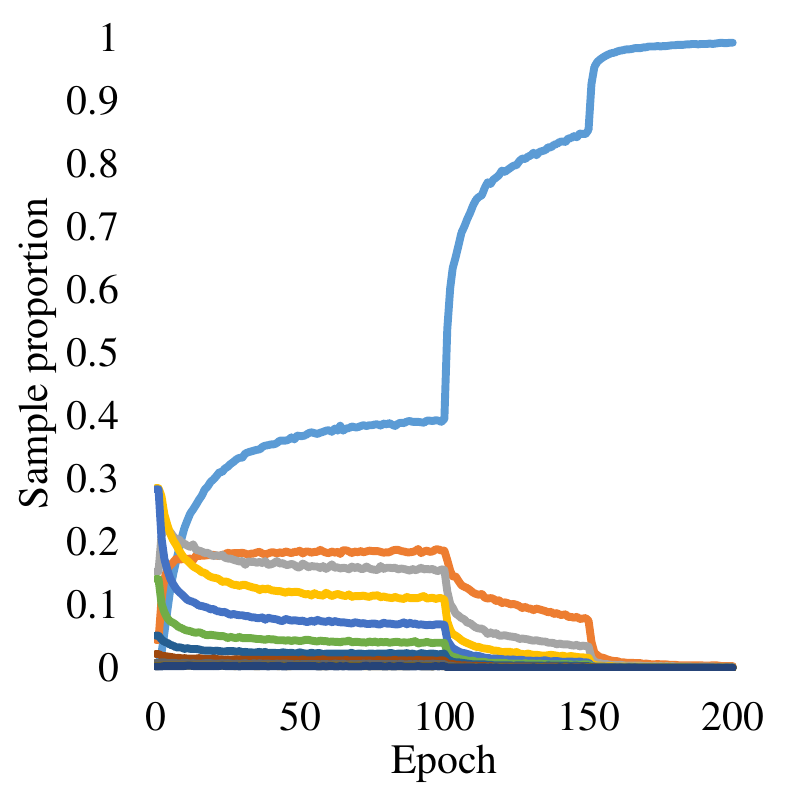}
        \includegraphics[width=0.24\columnwidth]{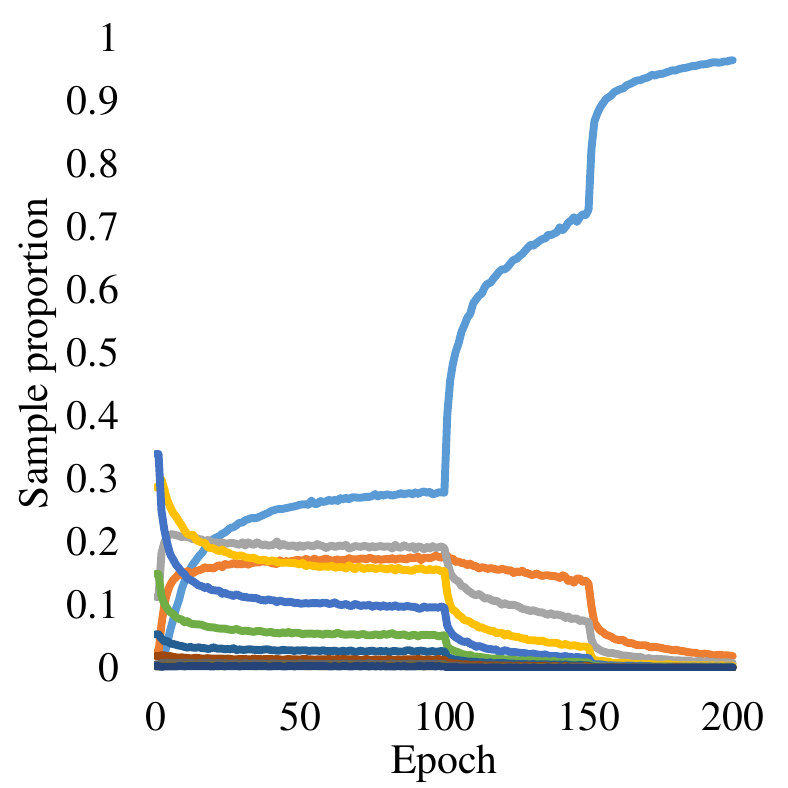}
        \includegraphics[width=0.24\columnwidth]{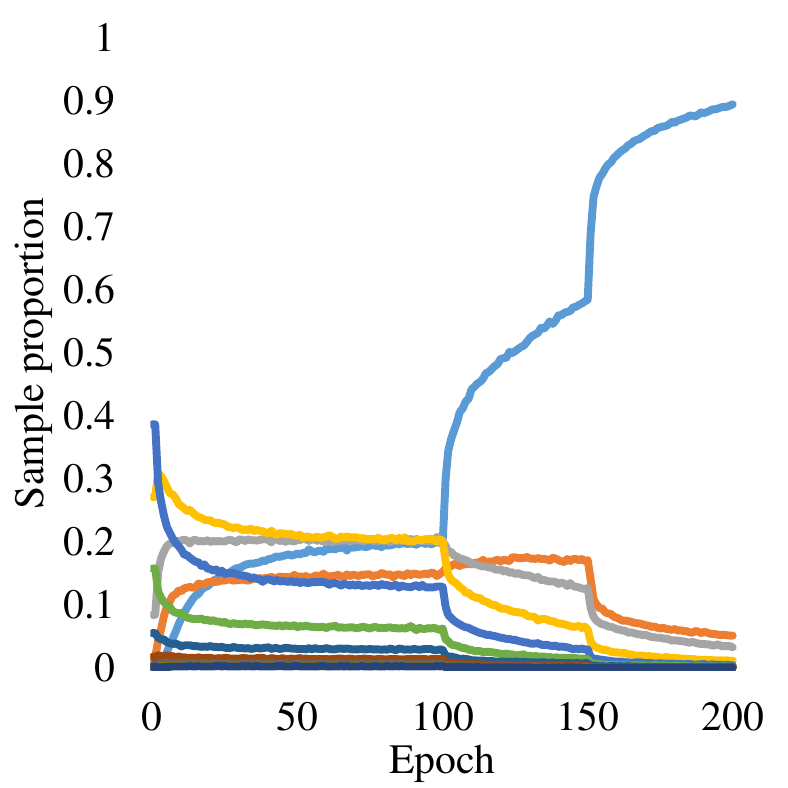}
    }
\caption{Robust overfitting behaviors and data distribution on CIFAR10 using Wide ResNet-34-10 under $L_\infty$ threat model. (a): The test robustness of adversarial training with various perturbation size $\epsilon$; (b) and (c): The distribution of training data in different loss ranges under various perturbation size $\epsilon$.}
\label{fig:7}
\end{figure}

In this section, we provide more empirical evidences for the robust overfitting behaviors and their data distributions across different datasets, model architectures and threat models. We use the same strategy in Section \ref{3-1} to adjust the strength of adversary. Specifically, for $L_\infty$ threat model, we vary $\epsilon$ from 0, 1, 2, 4, 8 to 10; for $L_2$ threat model, we vary $\epsilon$ from 0, 16, 32, 64, 128 to 160. As shown in Figure \ref{fig:4} to Figure \ref{fig:7}, we can always observe that there is no robust overfitting when the adversary is weak, and the robust overfitting phenomenon is particularly significant when the adversary is strong. Moreover, it can be seen that the data distribution of adversarial training with weak adversary mainly contains small-loss data, and the data distribution of adversarial training with strong adversary usually contains a considerable proportion of small-loss data and large-loss data. These evidences suggest that the observed robust overfitting behaviors and data distributions are general in adversarial training.

\section{More Evidences for the Causes of Robust Overfitting}
\label{AP_B}
\begin{figure}[h]
\centering
    \subfigure[CIFAR100 - PreAct ResNet-18 - $L_\infty$]{
        \includegraphics[width=0.24\columnwidth]{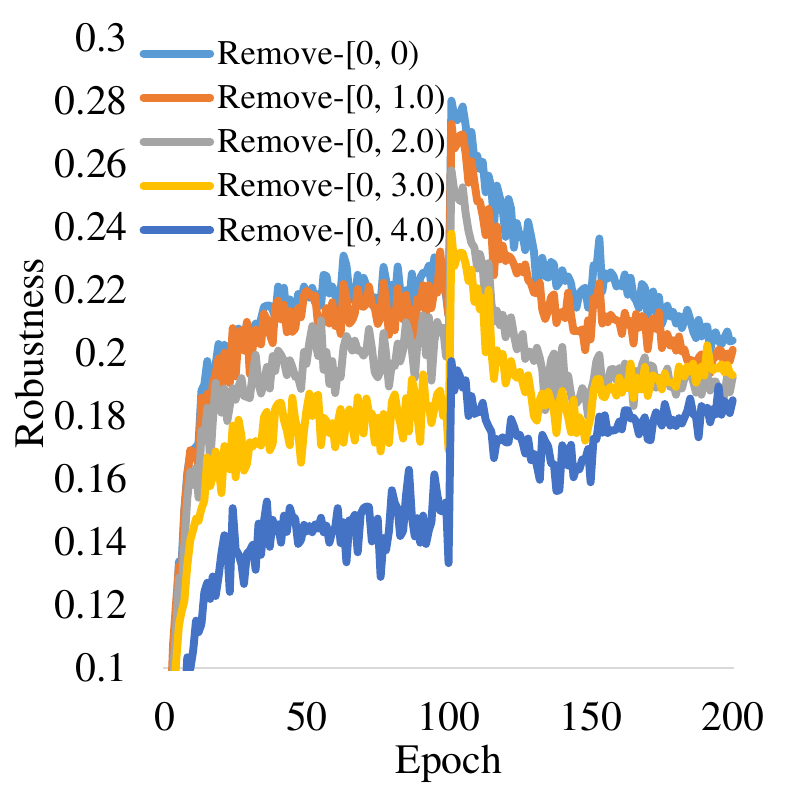}
        \includegraphics[width=0.24\columnwidth]{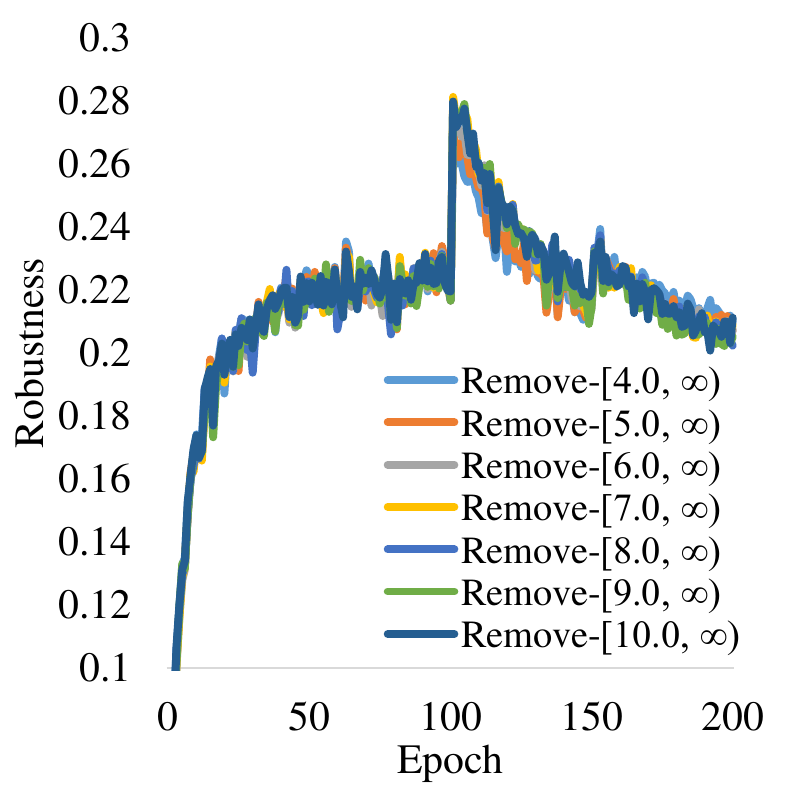}
    }
    \subfigure[SVHN - PreAct ResNet-18 - $L_\infty$]{
        \includegraphics[width=0.24\columnwidth]{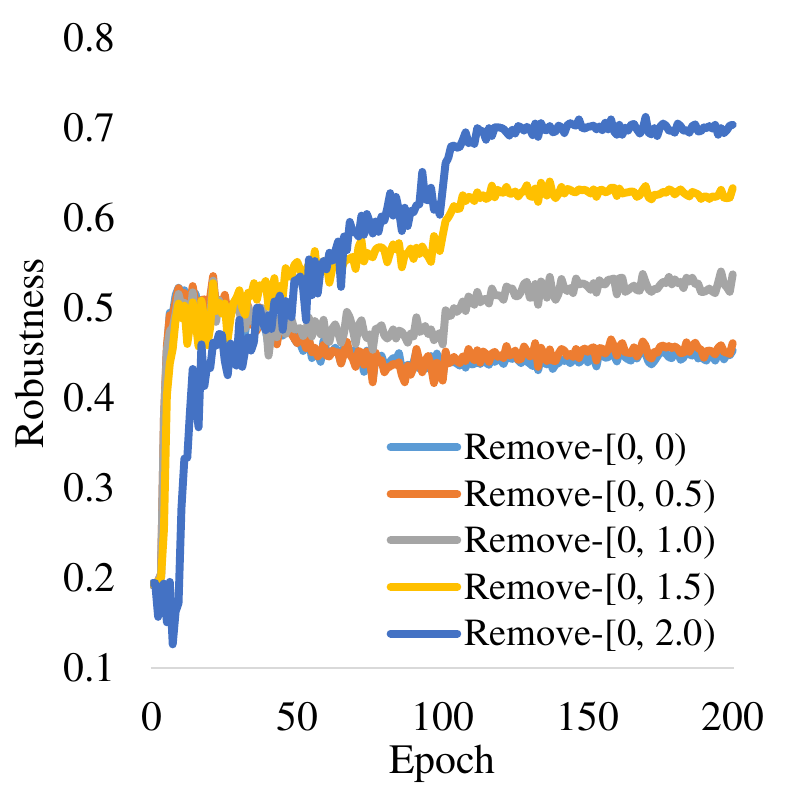}
        \includegraphics[width=0.24\columnwidth]{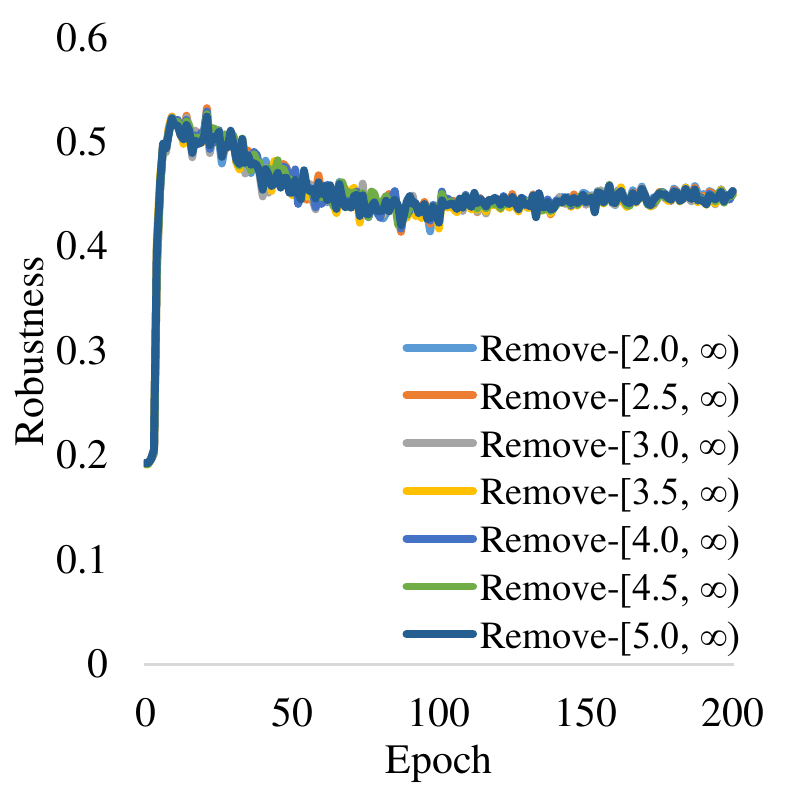}
    }
    \\
    \subfigure[CIFAR10 - PreAct ResNet-18 - $L_2$]{
        \includegraphics[width=0.24\columnwidth]{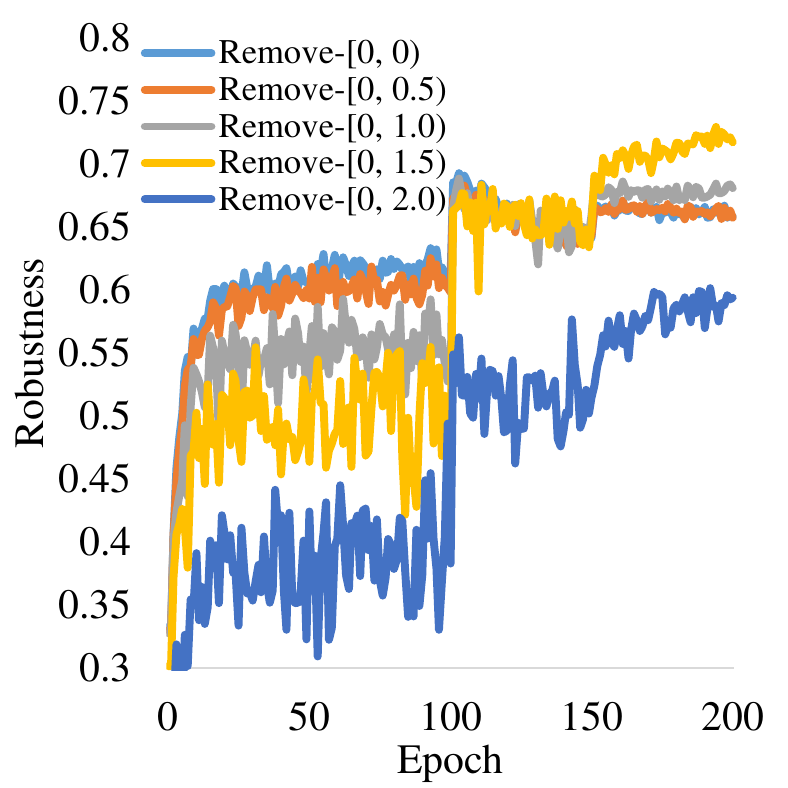}
        \includegraphics[width=0.24\columnwidth]{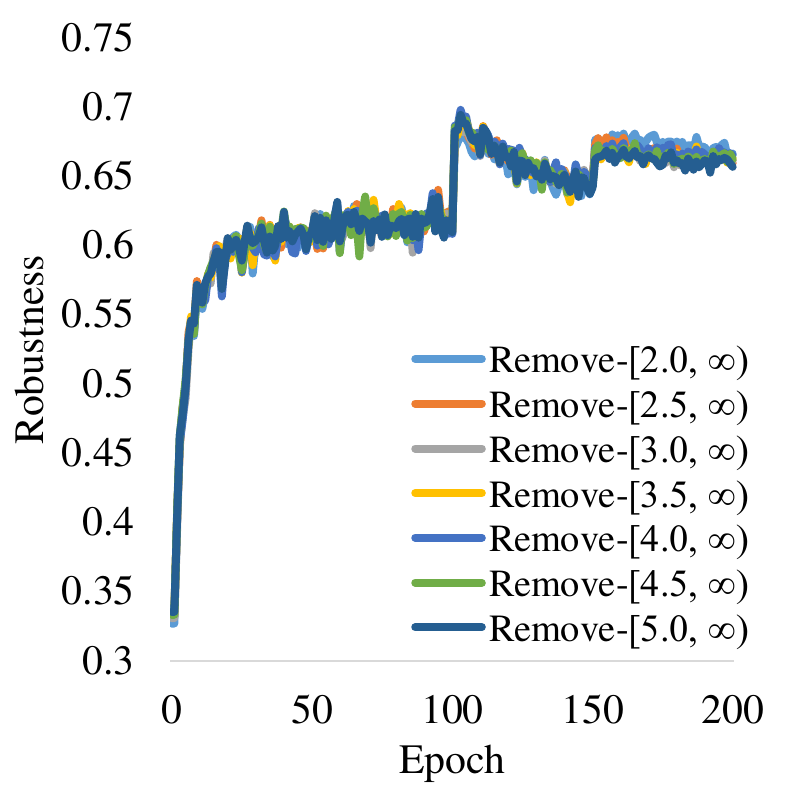}
    }
    \subfigure[CIFAR10 - Wide ResNet-34-10 - $L_\infty$]{
        \includegraphics[width=0.24\columnwidth]{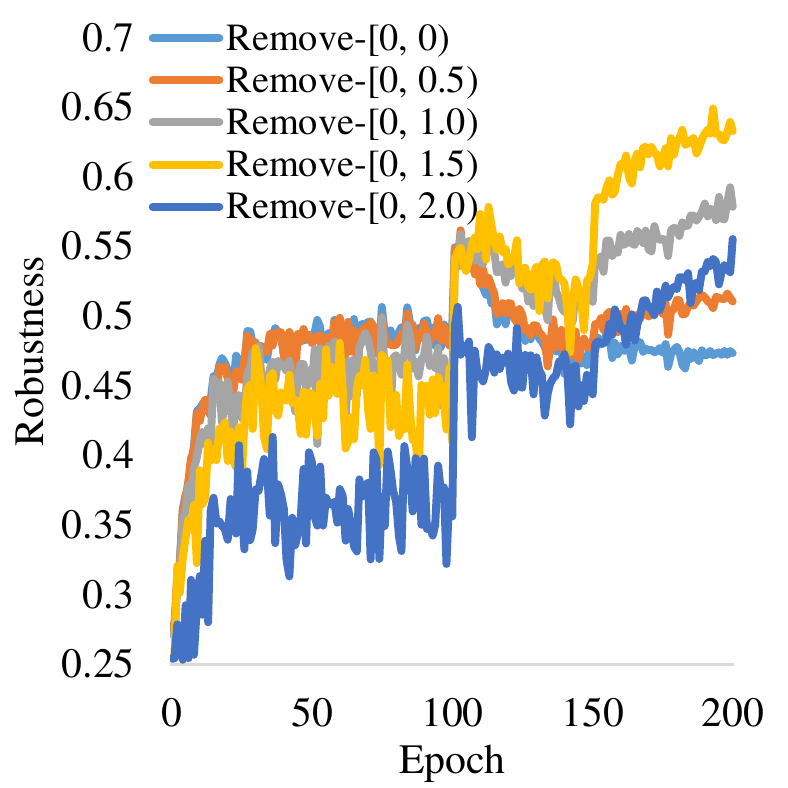}
        \includegraphics[width=0.24\columnwidth]{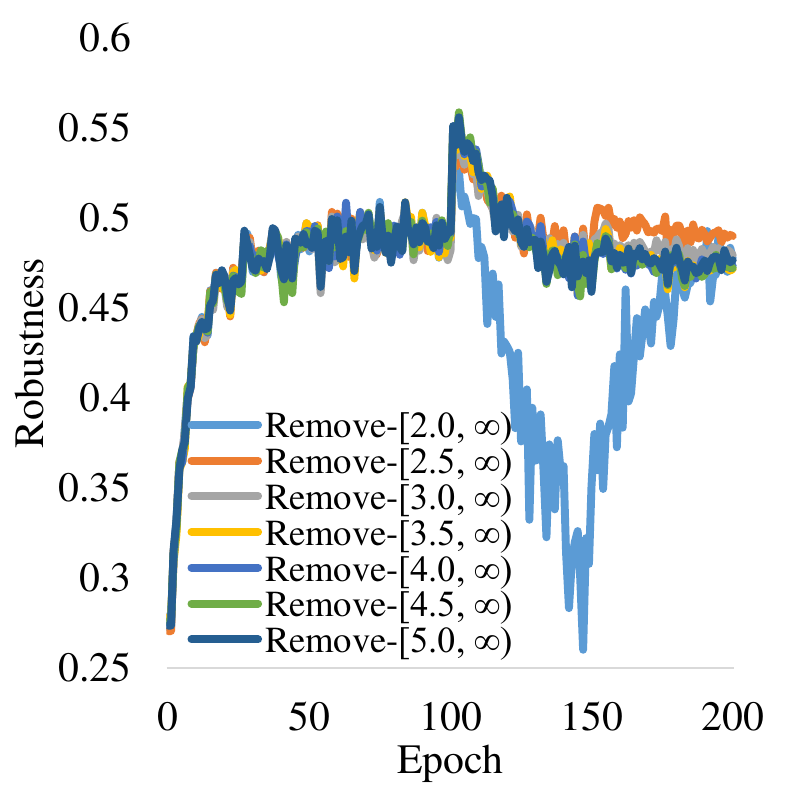}
    }
\caption{The data ablation experimental results on (a) CIFAR100 dataset using PreAct ResNet-18 under $L_\infty$ threat model; (b) SVHN dataset using PreAct ResNet-18 under $L_\infty$ threat model; (c) CIFAR10 dataset using PreAct ResNet-18 under $L_2$ threat model; (d) CIFAR10 dataset using Wide ResNet-34-10 under $L_\infty$ threat model.}
\label{fig:8}
\end{figure}

In this section, we further provide more evidences to verify that the small-loss data causes robust overfitting in strong adversary mode. We conduct data ablation adversarial training experiments across different datasets, network architectures and threat models. Specifically, we use $\epsilon=8$ for $L_\infty$ threat model and $\epsilon=128$ for $L_2$ threat model. We remove training data from various loss ranges during adversarial training. As shown in Figure \ref{fig:8}, robust overfitting phenomenon is basically unchanged after removing the large-loss data. However, the robust overfitting phenomenon can be eliminated after removing the small-loss data. These evidences clearly show that the robust overfitting in adversarial training is caused by these small-loss data.

\section{More Experimental Results}
\label{AP_C}
\subsection{Performance Evaluation}
\label{AP_C1}
In this part, we provide more performance evaluations of MLCAT$_{\mathrm{LS}}$ and MLCAT$_{\mathrm{WP}}$ on CIFAR10 dataset using PreAct ResNet-18 under $L_\infty$ threat model.

\textbf{Natural Accuracy.} The natural accuracy of AT, MLCAT$_{\mathrm{LS}}$ and MLCAT$_{\mathrm{WP}}$ are summarized in Table \ref{table:4}. It is observed that both MLCAT$_{\mathrm{LS}}$ and MLCAT$_{\mathrm{WP}}$ can achieve a fairly small performance gap between ``Best'' and ''Last'' on natural accuracy. Notably, MLCAT$_{\mathrm{WP}}$ is able to maintain the comparable natural accuracy to AT.

\textbf{Extension of MLCAT to TRADES.} We extend the proposed prototype to another well-recognized adversarial training variant TRADES. Specifically, for MLCAT-based TRADES (MLCTRADES), the inner maximization pass and outer minimization pass are in accordance with the TRADES method. In MLCTRADES$_{\mathrm{LS}}$ and MLCTRADES$_{\mathrm{WP}}$, we adopt the same $\ell_{min} = 1.5$ to distinguish small-loss data from large-loss data. As shown by the results in Table \ref{table:5}, it is evident that the proposed prototype significantly narrows robustness gap and MLCTRADES$_{\mathrm{WP}}$ outperforms the baseline method with a clear margin, demonstrating its effectiveness.

\textbf{Comparison with AWP.} Although both methods use the weight perturbation technique, MLCAT$_{\mathrm{WP}}$ and AWP are fundamentally different. First, their optimization objectives are different. MLCAT adopts an implicit adversarial example scheduling technique to eliminate robust overfitting, while AWP adopts the weight loss landscape.
Besides, the algorithm stability of MLCAT$_{\mathrm{WP}}$ is better than AWP. MLCAT$_\mathrm{WP}$ can work on both global and layer-wise perturbation scaling, while AWP suffers from training collapse on global perturbation scaling. Last but not least, we perform robustness comparison between MLCAT$_{\mathrm{WP}}$ and AWP, and the comparison results are summarized in Table \ref{table:6}. MLCAT$_{\mathrm{WP}}$ consistently outperforms AWP on all types of attacks, which fully demonstrates that our MLCAT can avoid robust overfitting and boost the robustness of adversarial training.

\begin{table}[h]
  \caption{Natural accuracy (\%) of AT, MLCAT$_{\mathrm{LS}}$ and MLCAT$_{\mathrm{WP}}$.}
  \label{table:4}
  \centering
  \begin{tabular}{lccc}
    \toprule
    \multirow{2}*{Method} & \multicolumn{3}{c}{Natural}\\
    \cmidrule{2-4}
    & Best & Last & Diff\\
    \midrule
    AT & 82.11 $\pm$ 0.45 & 84.72 $\pm$ 0.85  & 2.61 \\
    MLCAT$_{\mathrm{LS}}$ & 78.73 $\pm$ 0.77 & 79.48 $\pm$ 0.91 & 0.75\\
    MLCAT$_{\mathrm{WP}}$ & \textbf{84.1 $\pm$ 0.23} & \textbf{84.77 $\pm$ 0.35} & \textbf{0.67}\\
    \bottomrule
  \end{tabular}
\end{table}
\begin{table}[h]
  \caption{Test robustness (\%) under PGD-20 attack on TRADES.}
  \label{table:5}
  \centering
  \begin{tabular}{lccc}
    \toprule
    \multirow{2}*{Method} & \multicolumn{3}{c}{PGD20}\\
    \cmidrule{2-4}
    & Best & Last & Diff\\
    \midrule
    TRADES & 52.56 $\pm$ 0.43 & 49.12 $\pm$ 0.39 & -3.53 \\
    MLCTRADES$_{\mathrm{LS}}$ & 42.82 $\pm$ 0.25 & 41.4 $\pm$ 0.38 & -1.42\\
    MLCTRADES$_{\mathrm{WP}}$ & \textbf{55.28 $\pm$ 0.21} & \textbf{54.99 $\pm$ 0.19} & \textbf{-0.29}\\
    \bottomrule
  \end{tabular}
\end{table}

\begin{table}[h]
  \caption{Robustness comparison with AWP.}
  \label{table:6}
  \centering
  \begin{tabular}{lccccccc}
    \toprule
    \multirow{2}*{Method} & \multicolumn{3}{c}{PGD20} & &
    \multicolumn{3}{c}{AA}\\
    \cmidrule{2-4}
    \cmidrule{6-8}
    & Best & Last & Diff & &
    Best & Last & Diff\\
    \midrule
    AWP & 55.54 $\pm$ 0.20 & 54.64 $\pm$ 0.25 & -0.9 & & 49.94 $\pm$ 0.08 & 49.69 $\pm$ 0.10 & \textbf{-0.25}\\
    MLCAT$_{\mathrm{WP}}$ & \textbf{58.48 $\pm$ 0.39} & \textbf{57.65 $\pm$ 0.19} & \textbf{-0.83} & & \textbf{50.70 $\pm$ 0.11} & \textbf{50.32 $\pm$ 0.09} & -0.38\\
    \bottomrule
  \end{tabular}
\end{table}

\subsection{Ablation Studies}
\label{AP_C2}
\begin{figure}[h]
\centering
    \subfigure[CIFAR100]{
        \includegraphics[width=0.24\columnwidth]{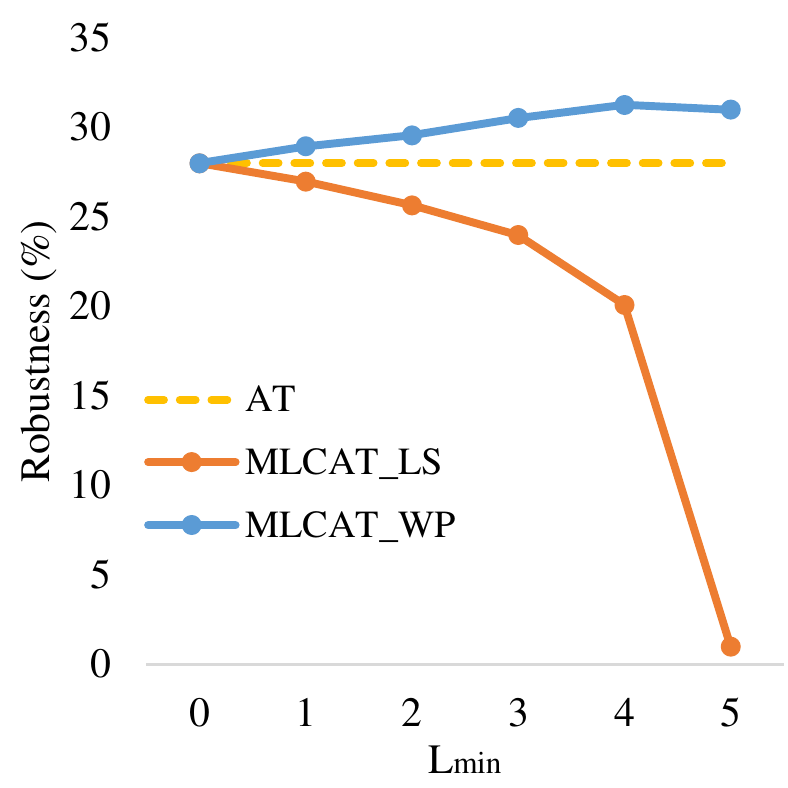}
        \includegraphics[width=0.24\columnwidth]{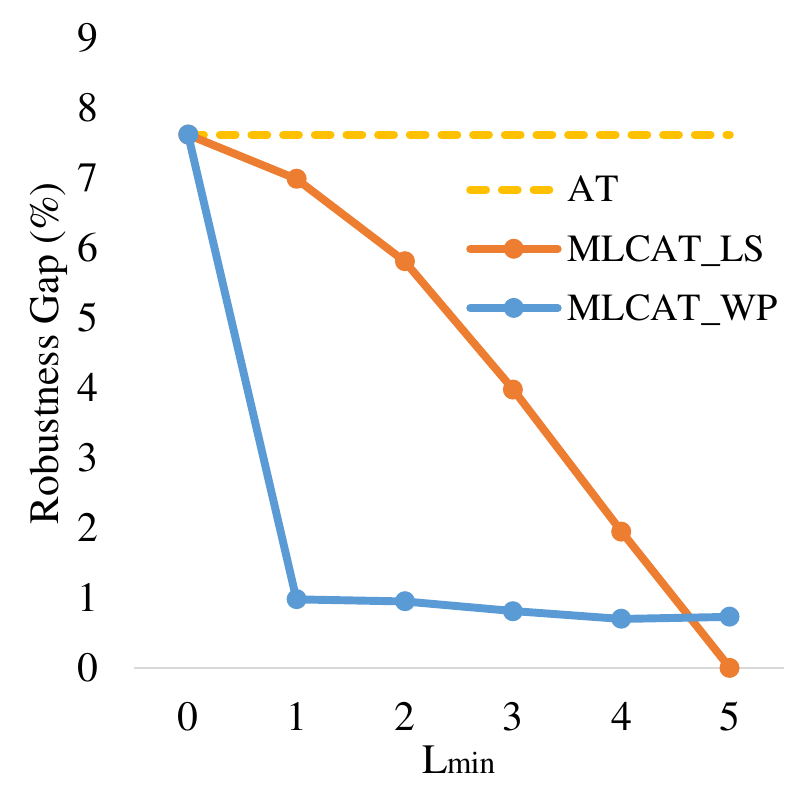}
    }
    \subfigure[SVHN]{
        \includegraphics[width=0.24\columnwidth]{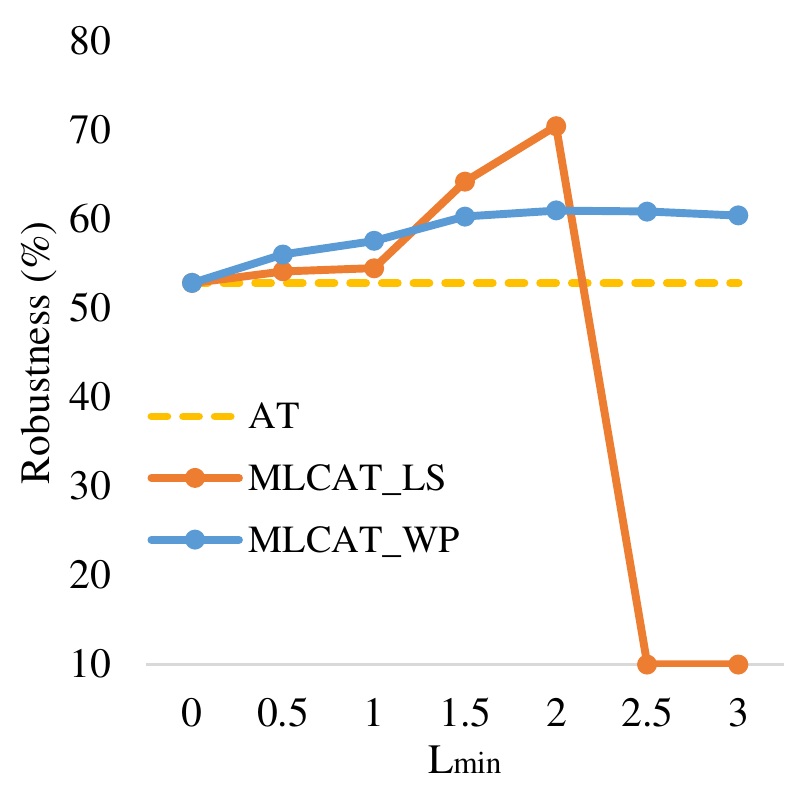}
        \includegraphics[width=0.24\columnwidth]{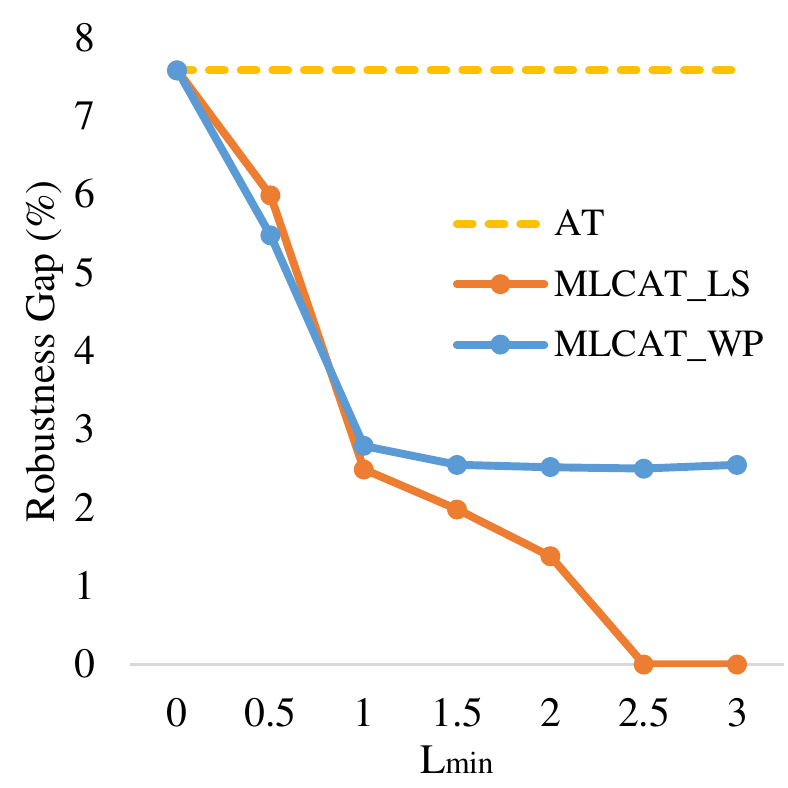}
    }
\caption{The experiment results of ablation study about minimum loss condition $\ell_{min}$ on (a) CIFAR100 dataset; (b) SVHN dataset.}
\label{fig:9}
\end{figure}

In this part, we provide the complete experimental results of ablation studies about the impact of minimum loss condition $\ell_{min}$ on CIFAR100 and SVHN datasets. Specifically, we vary the value of $\ell_{min}$ from 0 to 5.0 for CIFAR100, and from 0 to 3.0 for SVHN. The experimental results for the robustness performance and robustness gap are summarized in Figure \ref{fig:9}. It is observed that increasing $\ell_{min}$ consistently leads smaller robustness gap on CIFAR100 and SVHN datasets, and MLCAT$_{\mathrm{WP}}$ with a wide range of $\ell_{min}$ achieves better adversarial robustness than AT, demonstrating the importance of minimum loss condition $\ell_{min}$ in the MLCAT prototype.

\end{document}